\documentclass{article}

\PassOptionsToPackage{numbers,compress}{natbib}
\usepackage{natbib}

\usepackage[utf8]{inputenc}
\usepackage[T1]{fontenc}
\ifdefined\XeTeXversion
  \usepackage{newunicodechar}
  \newunicodechar{–}{--}
  \newunicodechar{—}{---}
  \newunicodechar{‘}{`}
  \newunicodechar{’}{'}
  \newunicodechar{“}{``}
  \newunicodechar{”}{''}
  \newunicodechar{ū}{\={u}}
  \newunicodechar{ė}{\.{e}}
  \newunicodechar{ą}{\k{a}}
  \newunicodechar{ę}{\k{e}}
  \newunicodechar{č}{\v{c}}
  \newunicodechar{š}{\v{s}}
\fi
\usepackage{hyperref}
\usepackage{arxiv}
\newcommand{\arxivfirstpageteaser}{}
\usepackage{authblk}
\makeatletter
\renewcommand\AB@affilsepx{,\hspace{1em}}
\makeatother
\usepackage{fancyhdr}
\fancypagestyle{firstpagestyle}{
  \fancyhf{}
  \renewcommand{\headrulewidth}{0pt}
}
\usepackage{url}
\DeclareRobustCommand{\modelname}[1]{{%
  \Urlmuskip=0mu\relax
  \def\UrlBigBreaks{\do\-}%
  \def\UrlBigBreakPenalty{0}%
  \def\UrlBreakPenalty{1000}%
  \nolinkurl{#1}%
}}
\usepackage{booktabs}
\usepackage{amsfonts}
\usepackage{nicefrac}
\usepackage{microtype}
\usepackage{xcolor}
\usepackage[normalem]{ulem}

\usepackage{amsmath,amssymb,amsthm,mathtools,bm}
\usepackage{graphicx}
\usepackage{listings}
\usepackage{tabularx}
\usepackage{multirow}
\usepackage{array}
\usepackage{float}
\usepackage{placeins}
\usepackage{needspace}
\usepackage{enumitem}
\usepackage{caption}
\usepackage{colortbl}
\usepackage{makecell}
\usepackage{wrapfig}
\usepackage{minitoc}
\usepackage{booktabs,longtable}
\usepackage{siunitx}
\usepackage[nameinlink,capitalize]{cleveref}

\newcommand{\bontaskcolumnwidth}{1.9cm}

\newcommand{\arxivwrappedlayout}{}
\AddToHook{env/wraptable/begin}{\captionsetup{font=small,skip=4pt}}
\AddToHook{env/wrapfigure/begin}{\captionsetup{font=small,skip=4pt}}

\newcommand{\reserveheadingroom}[1]{%
  \Needspace{#1}%
}

\AtBeginDocument{\raggedbottom}
\definecolor{rowgray}{gray}{0.92}
\definecolor{mayablue}{rgb}{0.21,0.49,0.74}
\definecolor{amaranth}{rgb}{0.9,0.17,0.31}
\definecolor{kwmagenta}{RGB}{200,0,160}
\definecolor{cmteal}{RGB}{0,140,140}
\definecolor{strorange}{RGB}{220,120,0}

\newif\ifshowrevisions
\showrevisionsfalse
\ifshowrevisions
  
  \long\def\del#1{{\color{red}\sout{#1}}\nobreak\hspace{0.25em}}
  
\else
  \usepackage{comment}
  
  \long\def\del#1{}
  \excludecomment{deletedblock}
\fi
\hypersetup{
  colorlinks=true,
  linkcolor=amaranth,
  citecolor=mayablue,
  urlcolor=teal,
  filecolor=magenta,
  breaklinks=true,
  bookmarksopen=true
}

\lstdefinestyle{pyalgo}{
  language=Python,
  basicstyle=\ttfamily\small,
  keywordstyle=\color{kwmagenta}\bfseries,
  commentstyle=\color{cmteal},
  stringstyle=\color{strorange},
  showstringspaces=false,
  columns=fullflexible,
  keepspaces=true,
  aboveskip=0.6em,
  belowskip=0.2em,
  breaklines=true,
  breakatwhitespace=true,
  breakindent=1.5em
}

\DeclareCaptionType{algorithm}[Algorithm][List of Algorithms]

\title{Next-token Preference: Scaling Unsupervised Reward Modeling}

\author[1]{Jingxuan Fan}
\author[2]{Yueying Li}
\author[1]{Zhenting Qi}
\author[3]{Dinghuai Zhang}
\author[1,4]{\\ Kiant\'e Brantley}
\author[1,4]{Sham M. Kakade}
\author[1]{Hanlin Zhang}
\affil[1]{Harvard University}
\affil[2]{Cornell University}
\affil[3]{Microsoft Research}
\affil[4]{Kempner Institute\noexpand\\\normalsize\textit{\today}}

\addtocontents{toc}{\protect\setcounter{tocdepth}{0}}

\begin{document}

\doparttoc
\faketableofcontents
\vspace*{-24pt}
\maketitle
\thispagestyle{firstpagestyle}

\begin{abstract}
Learning from feedback advances the capabilities and safety of frontier models, but preference supervision remains costly and difficult to scale.
We introduce next-token preference (NTP), an unsupervised reward-modeling method that constructs preference pairs from the continuation structure of raw web text.
NTP uses each prefix's original continuation as the chosen response and nearby continuations as in-batch negatives, without candidate generation or human or model-generated preference annotations.
Training on 11M tokens of math-focused web data yields steady gains on RewardBench and RewardBench 2, with gains across model families and sizes.
Across mathematical and general-domain web corpora, our method improves overall RewardBench 2 accuracy by 7.6--8.3 percentage points over initialization under the same token budget. Training on math-focused text yields gains of up to +16.1 on in-domain math subsets and consistent improvements on out-of-domain safety and general subsets.
These reward models improve best-of-$N$ selection on math, code, safety, and instruction following, and support math policy optimization competitive with supervised reward models of similar size.
NTP also improves the label efficiency of subsequent supervised fine-tuning. Using half the labeled pairs from a curated preference dataset, NTP initialization surpasses base initialization trained on the full dataset.
These results suggest that web text contains transferable preference signals that can reduce dependence on costly preference annotations.
\end{abstract}

\begingroup
\setlength{\intextsep}{8pt}
\ifdefined\arxivfirstpageteaser
\begin{figure}[H]
\else
\begin{figure*}[ht]
\fi
    \centering
    \includegraphics[width=0.83\linewidth]{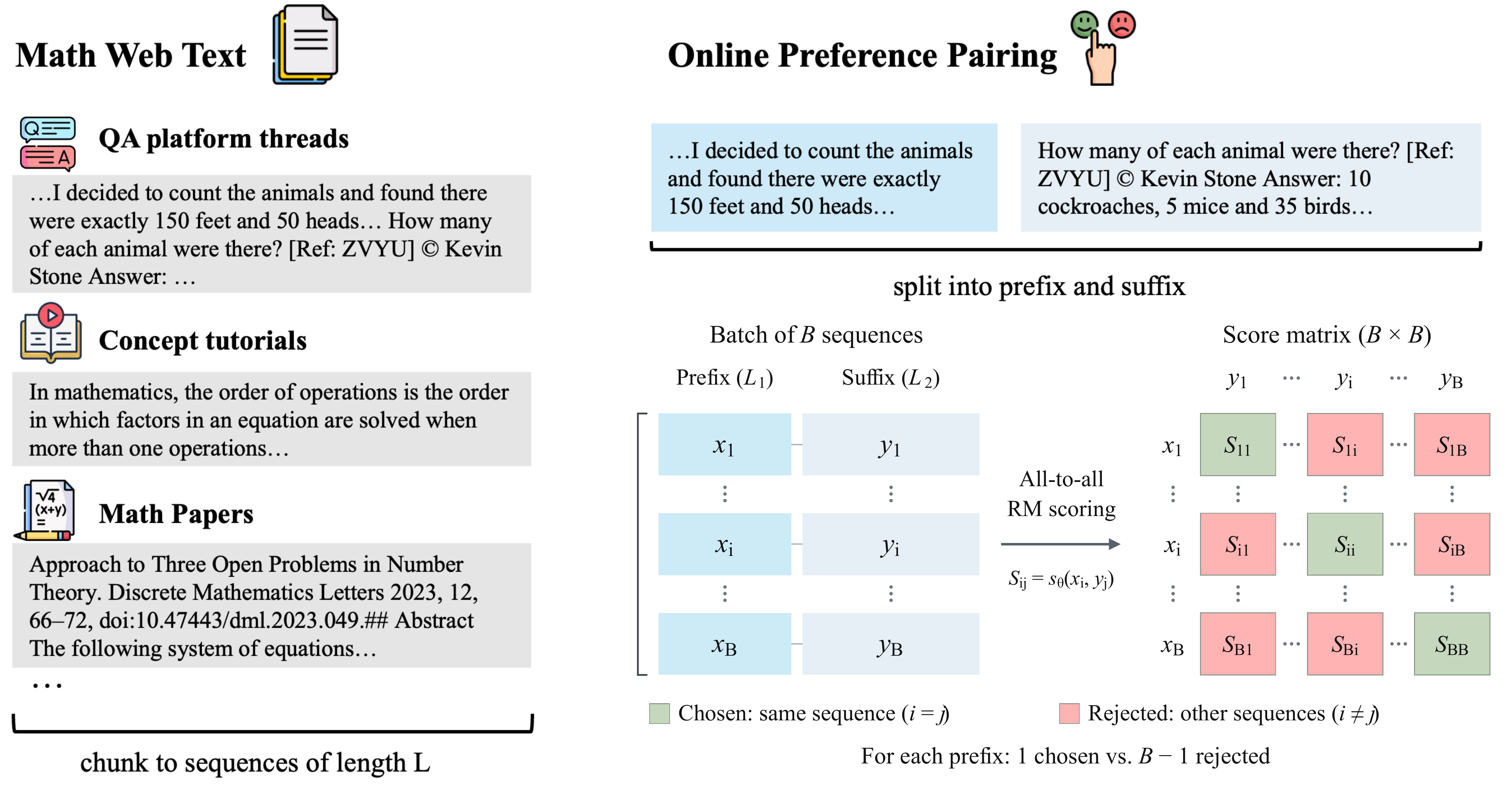}
    \caption{Schematic overview of our NTP reward model training workflow from web math text.
    Raw documents are split into prefix–suffix pairs, where the true continuation is treated as the chosen response and other in-batch continuations serve as implicit negatives. The reward model is trained with a Bradley–Terry objective over these online preference pairs, enabling scalable reward learning without human annotations.}%
    \label{fig:teaser}
\vskip -0.1in
\ifdefined\arxivfirstpageteaser
\end{figure}
\else
\end{figure*}
\fi

\endgroup
\clearpage
\ifdefined\arxivfirstpageteaser\else
\ifdefined\arxivfirstpageteaser
\begin{figure}[H]
\else
\begin{figure*}[ht]
\fi
    \centering
    \includegraphics[width=0.83\linewidth]{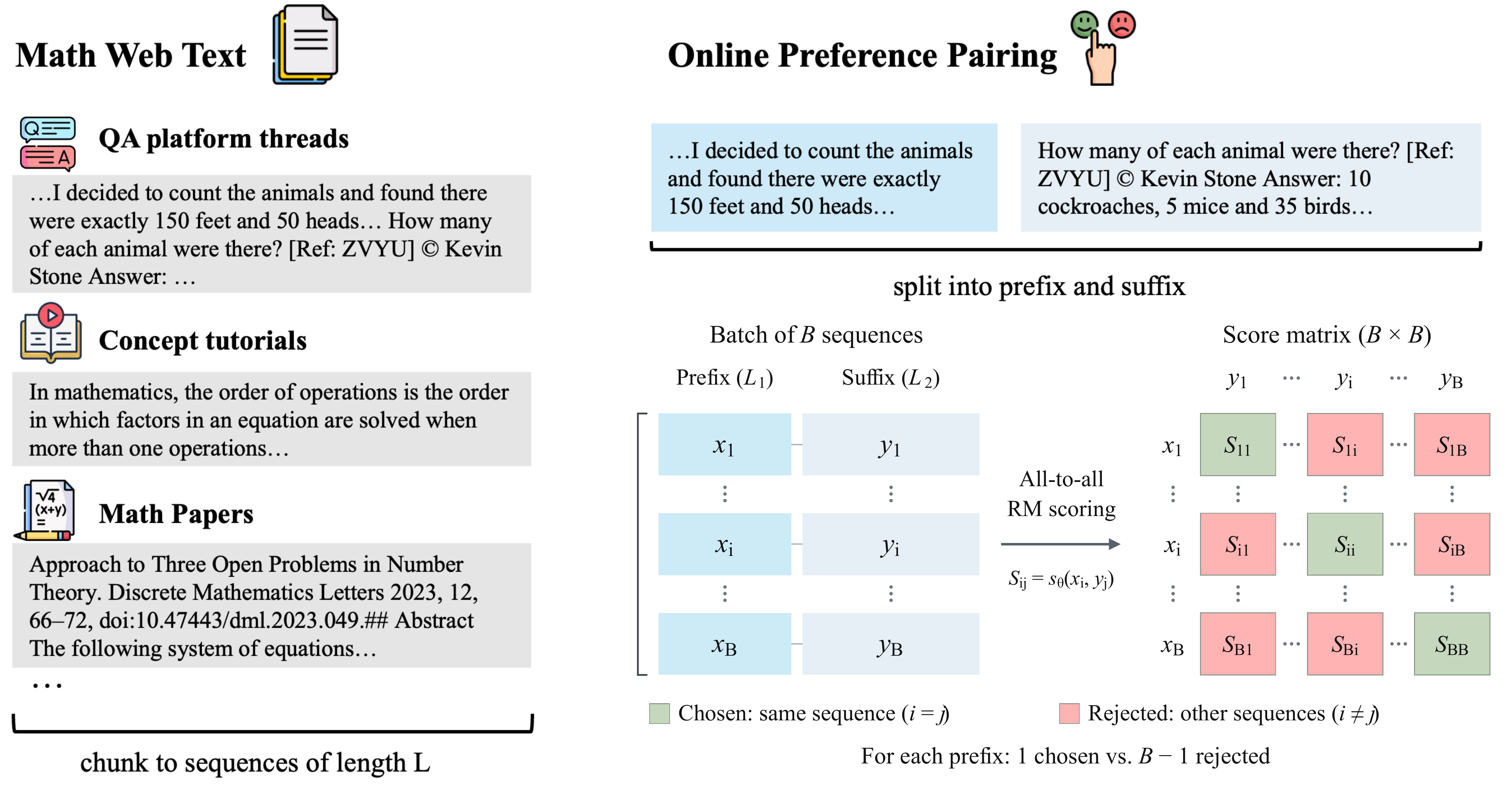}
    \caption{Schematic overview of our NTP reward model training workflow from web math text.
    Raw documents are split into prefix–suffix pairs, where the true continuation is treated as the chosen response and other in-batch continuations serve as implicit negatives. The reward model is trained with a Bradley–Terry objective over these online preference pairs, enabling scalable reward learning without human annotations.}%
    \label{fig:teaser}
\vskip -0.1in
\ifdefined\arxivfirstpageteaser
\end{figure}
\else
\end{figure*}
\fi

\fi
\section{Introduction}
\label{sec:intro}

Reinforcement learning from human feedback (RLHF) \citep{christiano2017deep} has been the workhorse for building helpful and harmless language models \citep{ouyang2022training, bai2022training}. Reward models support this goal by guiding response selection and policy optimization.
Their effectiveness depends on the preference data used to train them, yet obtaining these data at scale remains costly.
Conventional pipelines require candidate generation and preference judgments from human annotators or models \citep{cui2024ultrafeedbackboostinglanguagemodels, bai2022training}; automating judgments still incurs generation and scoring costs (\cref{sec:ex_cost}). Moreover, human feedback can inherently be noisy in part due to annotator subjectivity, inconsistency, and labeling errors, a phenomenon widely observed in preference-based learning \mbox{\citep{christiano2017deep, ouyang2022training, casper2023open, zhang2024diverging, gao2024impact}}. 
Such systematic noise can significantly misguide reward models and produce serious misalignment downstream \mbox{\citep{korbak2023pretraining, macdiarmid2025natural}}. Furthermore, as model capabilities scale, we face aligning systems whose relevant behaviors and failure modes exceed what humans (and comparably weak models) can reliably judge, making human-labeled reward signals fundamentally limited \mbox{\citep{bowman2022measuringprogressscalableoversight,engels2025scaling}}. These constraints motivate sources of reward-model supervision that do not require an explicit judgment for every comparison.

To address this supervision bottleneck, we introduce next-token preference (NTP), an unsupervised reward-modeling method that constructs preference pairs from the continuation structure of raw web text, without human annotations or model-generated judgments.
We split each text chunk into a prefix and continuation, using the prefix as the prompt and its original continuation as the chosen response (\cref{fig:teaser,sec:methods}).
Continuations from nearby chunks serve as rejected responses through in-batch pairings \citep{hadsell2006dimensionality, chen2020simple}.
The observed relationship between a prefix and its continuation thus supplies an implicit preference signal.
Nearby continuations can share topic, vocabulary, and style while breaking the dependencies that connect a prefix to its original continuation.
Comparing them encourages the reward model to learn semantic and logical consistency across the prefix--continuation boundary.
All comparisons are assembled online from existing text, allowing preference supervision to scale with the corpus without candidate generation or preference annotation.

We use this construction to investigate whether the structure of web text can supply the supervision needed to learn useful, generalizable reward functions. Our main contributions are:
\begin{itemize}[leftmargin=*,topsep=4pt,itemsep=3pt,parsep=0pt]
    \item \textbf{Transferable reward learning from web text.}
    Continuation consistency supports broader judgments of response quality: models trained on math text improve both reasoning and out-of-domain safety without labels for these behaviors.
    Gains grow with training data and extend across model families and sizes.
    Across two mathematical corpora and general-domain \textit{FineWeb-Edu}, only 11M tokens per corpus yield peak overall RewardBench 2 gains of \textbf{7.6--8.3} percentage points over initialization \citep{malik2025rewardbench2advancingreward}.
    \item \textbf{Downstream utility.}
    The learned rewards guide improvements in language-model behavior through both response selection and policy optimization.
    They improve best-of-$N$ selection on math, code, safety, and instruction following, and support mathematical policy optimization competitive with supervised reward models of comparable size.
    \item \textbf{A foundation for label-efficient reward modeling.}
    NTP also complements curated preference supervision by providing an effective mid-training initialization.
    With only half of the labeled \texttt{RLHFMix1} data, NTP initialization surpasses base initialization trained on the full dataset, supporting unsupervised reward learning as a scalable precursor to supervised training.
\end{itemize}

\section[Continuation-Based Preference-Pair Construction]{Continuation-Based Preference-Pair Construction}
\label{sec:methods}
\label{sec:algorithm}

A reward model assigns a scalar score $s_{\theta}(x,y)$ to a response $y$ conditioned on a prompt $x$ and is trained to rank a preferred response $y^{+}$ above a rejected response $y^{-}$. Standard pipelines require human-written or model-generated prompts and candidate responses, together with preference annotations, all of which incur substantial costs at scale (see \cref{sec:ex_cost}). We introduce next-token preference (NTP), an unsupervised method that trains reward models directly from raw web text without relying on curated preference pairs. NTP uses a text prefix as the prompt and its original suffix as the chosen response. Rejected responses are suffixes from nearby text chunks that share the surrounding context but do not directly follow that prefix. We form these comparisons by grouping nearby chunks into a batch and pairing each prefix with the other chunks' suffixes. This provides preference supervision without generating new responses or collecting preference judgments.

\begin{wrapfloat}{algorithm}{r}{0.53\textwidth}
\vspace{-\intextsep}
\captionsetup{type=algorithm}
\caption[NTP: Reward Model Training from Web Data]{NTP: Reward Model Training from Web Data}
\label{alg:online_bt}
\begin{lstlisting}[
  style=pyalgo,
  escapeinside={(*@}{@*)},
  basicstyle=\ttfamily\footnotesize,
  breaklines=true,
  breakatwhitespace=true,
  columns=fullflexible,
  keepspaces=true,
  aboveskip=0.3em,
  belowskip=0pt
]
# split(): breakpoint sampler on a continuous text sequence
# RM(): reward model that outputs a scalar score for (prompt, response)
for seq_batch in data_loader:
    prompts, responses = split(seq_batch)
    # S[i, j] = RM(x_i, y_j)
    score_mat = RM(prompts, responses)
    # S, [B, B]
    pos_scores = diag(score_mat)
    # s_pos, [B] (self-pairs = chosen)
    neg_scores = offdiag(score_mat)
    # s_neg, [B, B-1] (cross-pairs = rejected)
    bt_loss = BT(pos_scores, neg_scores)
    c_loss  = CENTER(pos_scores, neg_scores)
    loss = bt_loss + c_loss
    loss.backward()
    update(RM.param)

def BT(s_pos, s_neg):
    # Bradley-Terry loss
    logits = s_pos[:, None] - s_neg
    return -log_sigmoid(logits).mean()

def CENTER(s_pos, s_neg):
    # score-centering regularizer
    return centering_coeff * (s_pos**2 + s_neg**2).mean()
\end{lstlisting}

\end{wrapfloat}
In our main experiments, we construct comparisons from \textit{FineMath-4plus} \citep{allal2025smollm2smolgoesbig}, a collection of mathematical web documents filtered from CommonCrawl. We additionally use \textit{MegaMath-Web-Pro-Max} \citep{wang2025octothinkermidtrainingincentivizesreinforcement}, another mathematical web corpus, and \textit{FineWeb-Edu} \citep{penedo2024fineweb}, a general-domain educational web corpus, to test generalization to other math and non-math web corpora (contamination checks in \cref{sec:ex_contamination}).

For each corpus, we concatenate documents into a continuous token stream and divide it into fixed-length blocks of $L=L_1+L_2$ tokens. Each block is split into an $L_1$-token prefix $x$ and an $L_2$-token continuation $y$, forming a prompt--response pair $(x,y)$ (\cref{fig:teaser}). The split need not coincide with a sentence boundary (dataset parameters and sensitivity to split position in \cref{sec:ex_dataset_params}).

Grouping $B$ nearby chunks from this stream gives a batch of prefix--continuation pairs $\{(x_i,y_i)\}_{i=1}^{B}$. For each prefix $x_i$, we set $y_i^{+}=y_i$ and $y_{ij}^{-}=y_j$ for every $j\neq i$. Each prompt therefore yields $B-1$ comparisons of the form
\[
(x_i,y_i^{+},y_{ij}^{-})=(x_i,y_i,y_j),\qquad j\neq i.
\]
This all-to-all construction yields $B(B-1)$ preference comparisons per batch, assembled online during training. Suffixes from nearby chunks of the same mathematical article can share topic, vocabulary, and style while breaking the dependencies that connect a prefix to its true suffix. Their shared context limits shortcuts based on topic or style and encourages the reward model to learn semantic and logical dependencies across the prefix--suffix boundary. Negative difficulty depends on batch composition and splitting (\cref{sec:ablation,sec:ex_proxy_judge}). Our experiments on each representative corpus use only 11M tokens, a small fraction of the trillions of tokens available in web corpora \citep{penedo2024fineweb}. Applying the same construction to additional web text expands the supply of preference comparisons without candidate generation or preference annotation.

We train the reward model on these comparisons with the Bradley--Terry objective \citep{19ff28b9-64f9-3656-ba40-08326a05748e}, averaging the negative log-likelihood over all $B(B-1)$ in-batch comparisons:
\begin{equation}
\label{eq:bt_loss}
\mathcal{L}_{\text{BT}}
= -\frac{1}{B(B-1)}\sum_{i=1}^{B}\sum_{j\neq i}
\log \sigma\!\big(s_{\theta}(x_i,y_i^{+})-s_{\theta}(x_i,y_{ij}^{-})\big).
\end{equation}
To stabilize training, we add score-centering regularization to the Bradley--Terry objective \citep{eisenstein2024helpingherdingrewardmodel}. This is motivated by two properties of Bradley--Terry training: (i) the preference likelihood depends only on score differences and is therefore underdetermined up to a prompt-dependent offset, allowing the absolute reward scale to drift; and (ii) unconstrained optimization can inflate score magnitudes to produce overconfident margins that amplify spurious correlations and induce heavy-tailed reward distributions that are harmful for downstream selection. We therefore penalize large-magnitude reward outputs, encouraging a well-behaved and comparable score scale across prompts:
\begin{equation}
\label{eq:center_loss}
\mathcal{L}_{\text{center}} = \mathbb{E}\Big[s_{\theta}(x_i,y_i^{+})^2 + \frac{1}{B-1}\sum_{j\neq i}s_{\theta}(x_i,y_{ij}^{-})^2\Big].
\end{equation}

Our final RM training loss is
\begin{equation}
\label{eq:total_loss}
\mathcal{L}=\mathcal{L}_{\text{BT}}+c\,\mathcal{L}_{\text{center}},
\end{equation}
where $c$ is the centering coefficient. Constraining the reward range reduces overfitting to corpus-specific artifacts and improves robustness across diverse prompts and continuations (\cref{alg:online_bt}).

\section[Results]{Results}
\label{sec:exp}

\subsection[Reward-Model Training and Preference-Alignment Evaluation]{Reward-Model Training and Preference-Alignment Evaluation}
\label{sec:preference_alignment}
\label{sec:experimental_setup}
\label{sec:generalization}

We evaluate NTP by training reward models from four backbones spanning two model families and parameter scales from 1B to 7B: base \modelname{Llama-3.2-1B} and \modelname{Llama-3.2-3B} \citep{grattafiori2024llama3herdmodels}, and instruction-tuned \modelname{Qwen2.5-3B-Instruct} and \modelname{Qwen2.5-7B-Instruct} \citep{qwen2025qwen25technicalreport} (recipe and hyperparameters in \cref{sec:ex_rm_hparams,tab:rm_hparams}). All four models are trained on the web-text comparisons constructed in \cref{sec:methods} using objective in \cref{eq:total_loss}, under the same 11M-token budget. We use \modelname{Llama-3.2-3B} as the representative backbone for controlled ablations.

We first evaluate the trained reward models' general preference alignment. We use RewardBench \citep{lambert2024rewardbenchevaluatingrewardmodels} and RewardBench 2 \citep{malik2025rewardbench2advancingreward}, two benchmarks that measure whether reward models rank preferred responses above rejected alternatives across several dimensions of response quality. RewardBench covers response helpfulness and instruction following (\texttt{Chat}, \texttt{Chat Hard}), refusal of harmful requests while answering benign ones (\texttt{Safety}), and mathematical and coding correctness (\texttt{Reasoning}). RewardBench 2 covers mathematical correctness (\texttt{Math}), appropriate responses to harmful and benign but potentially risky requests (\texttt{Safety}), sensitivity to factual errors (\texttt{Factuality}), reliable ranking of responses with similar quality (\texttt{Ties}), relevance to the requested task (\texttt{Focus}), and compliance with verifiable instructions (\texttt{Precise IF}). Because our primary corpus is mathematical, we report \emph{in-domain} (ID)
performance (\texttt{Reasoning} subset for RewardBench; \texttt{Math} subset for RewardBench 2), out-of-domain (OOD) performance on safety (\texttt{Safety} subsets for both) and other remaining subsets, alongside the overall
benchmark average.%

\begin{figure*}[t]
  \centering
  \includegraphics[width=0.95\textwidth]{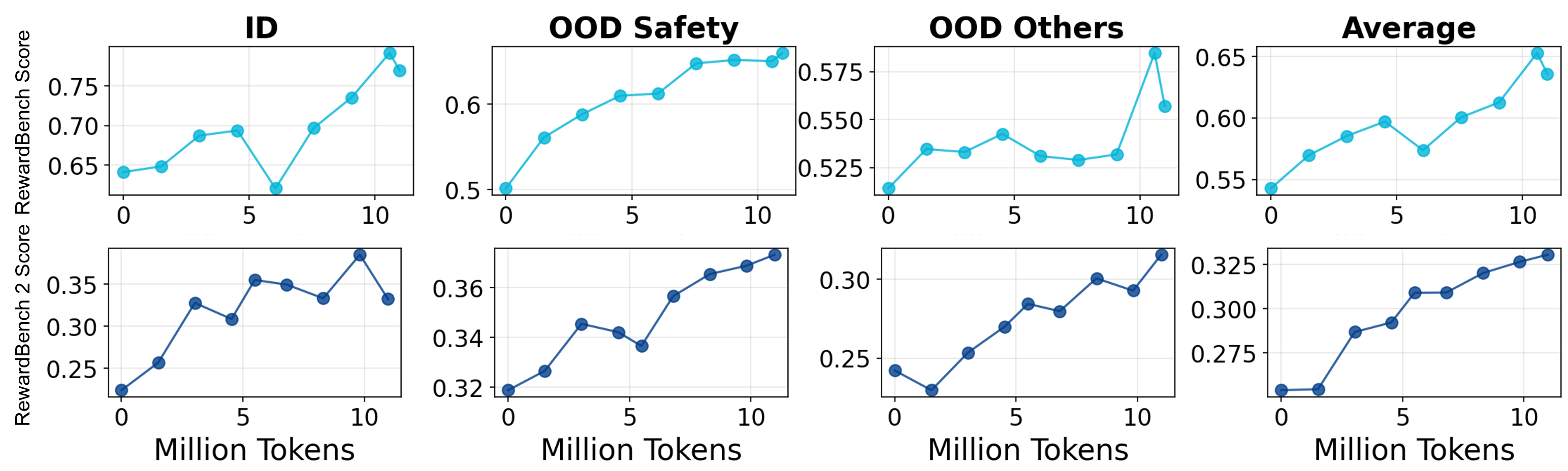}
  \caption{Scalability of NTP with respect to data size. Reward models initialized from pretrained Llama-3.2-3B improve across RewardBench and RewardBench 2 subsets as the token budget increases to 11M.}
  \label{fig:3b_rb1_rb2}
\vskip -0.1in
\end{figure*}

\noindent For \modelname{Llama-3.2-3B}, RewardBench and RewardBench 2 accuracy improves steadily over the 11M-token budget (\cref{fig:3b_rb1_rb2,sec:ex_multiseed}), suggesting that large-scale raw text provides a useful preference-learning signal even without curated comparisons. At 11M tokens, the gains extend beyond mathematical reasoning to general preference alignment on both benchmarks. On RewardBench, accuracy in judging appropriate refusal improves by 16.0 percentage points over initialization, exceeding the 13.1-point gain in evaluating mathematical and coding correctness. Accuracy in identifying better responses in challenging conversations also improves by 7.0 points. RewardBench 2 likewise shows substantial improvements in judging response safety and relevance to the user's request (\cref{tab:rb_subset_gains}). To understand this surprising transfer from mathematical text to safety, we examine performance on individual questions in the RewardBench 2 \texttt{Safety} subset. The analysis indicates that the gains extend beyond learning superficial refusal cues (\cref{sec:ex_safety_transfer}). Thus, although training only contrasts consistent mathematical continuations with mismatched ones, the learned reward function transfers to judgments of appropriate refusal, response helpfulness, and task relevance without explicit preference supervision for these behaviors.\par
\Needspace{14\baselineskip}
\begin{wraptable}{r}{0.48\textwidth}
\vspace{-0.6\baselineskip}
\normalcolor
\centering
\caption{Overall RewardBench 2 gains (percentage points) over \modelname{Llama-3.2-3B} initialization. \textbf{Bold}: best gain at the matched 11M-token budget.}
\label{tab:curated_comparison}
\scriptsize
\setlength{\tabcolsep}{3pt}
\renewcommand{\arraystretch}{1.1}
\begin{tabular*}{\linewidth}{@{\extracolsep{\fill}}llr@{}}
\toprule
\textbf{Method} & \textbf{Budget} & \textbf{RewardBench 2 gain} \\
\midrule
FineMath NTP (ours)& 11M tokens & \textbf{+7.7} \\
\midrule
Skywork-80K-v0.2 & $\approx$9K pairs (11M) & $+1.8$ \\
Skywork-80K-v0.2 & 77K pairs (full) & $+9.8$ \\
\midrule
RLHFMix1 & $\approx$22K pairs (11M) & $+0.8$ \\
RLHFMix1 & 244K pairs (full) & $+8.3$ \\
\bottomrule
\end{tabular*}
\vspace{-0.5\baselineskip}
\end{wraptable}

\noindent Using RewardBench 2, we also compare against training the same \modelname{Llama-3.2-3B} backbone under a matched 11M-token budget on \texttt{Skywork-80K-v0.2}, a curated dataset of approximately 77K labeled preference pairs \citep{liu2024skyworkrewardbagtricks}, and \texttt{RLHFMix1}, a curated mixture of 244K preference pairs labeled by humans and GPT-4 \citep{dong2024rlhf}. At the same token budget, NTP improves overall accuracy by 7.7 percentage points, compared with 1.8 for Skywork and 0.8 for RLHFMix1 (\cref{tab:curated_comparison}). This gain also approaches those from training on the full curated datasets (9.8 and 8.3 points, respectively), despite using raw web text without human or LLM preference labels. Dataset construction and equivalent pair matching are detailed in \cref{sec:ex_curated_datasets_stats}.\par
\WFclear

\noindent The same training recipe also transfers across the other backbones (\cref{tab:rb_averages,sec:ex_rm_ranking}). Competitive performance extends to the largest backbone tested. The reward model initialized from \modelname{Qwen2.5-7B-Instruct} ranks in the upper half of the reported 7B comparison groups on both RewardBench and RewardBench 2. These results suggest that NTP remains competitive as model size grows and support its applicability across model families and base and instruction-tuned initializations.\par

\subsection[Ablation Studies]{Ablation Studies}
\label{sec:ablation}
\label{sec:ablation_controls}

Using \modelname{Llama-3.2-3B} and the default training setup above, we ablate batch size, data splitting, and score centering loss. We also compare against continued pretraining on the same text to test whether corpus exposure alone explains the gains. 

\ifdefined\arxivwrappedlayout
\begin{figure*}[!htbp]
\else
\begin{figure*}
\fi
\centering

\begin{minipage}[t]{0.34\textwidth}\vspace{0pt}
  \centering
  \small
  \setlength{\tabcolsep}{2pt}
  \renewcommand{\arraystretch}{1.05}
\vskip 0.2in
  \resizebox{\linewidth}{!}{%
    \begin{tabular}{lrrrr}
      \toprule
      \textbf{batch size} &
      \makecell{\textbf{ID}} &
      \makecell{\textbf{OOD}\\\textbf{Safety}} &
      \makecell{\textbf{OOD}\\\textbf{others}} &
      \makecell{\textbf{RewardBench 2}} \\
      \midrule
      8  & +9.6  & +3.4          & +0.6          & +1.0 \\
      16 & +10.7 & \textbf{+8.3} & \textbf{+7.7} & +6.7 \\
      \rowcolor{rowgray}
      32 & \textbf{+16.1} & +5.4 & +7.4 & \textbf{+7.7} \\
      \bottomrule
    \end{tabular}%
  }

  \vspace{2pt}
  {\footnotesize (a) Peak RewardBench 2 subsets and average gains vs. batch size}
\end{minipage}\hfill
\begin{minipage}[t]{0.64\textwidth}\vspace{0pt}
  \centering
  \includegraphics[width=\linewidth]{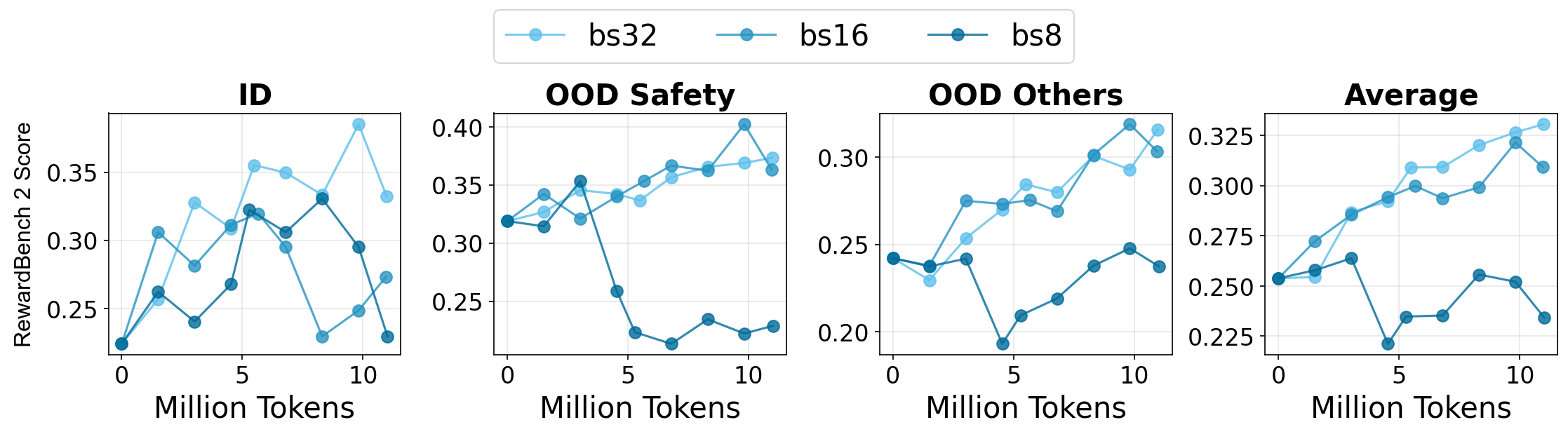}

  \vspace{1pt}
  {\footnotesize (b) RewardBench 2 learning curves across 11M token budget vs. batch size}
\end{minipage}

\caption{Effect of batch size on peak gains and learning trajectory of RewardBench 2.}
\label{fig:bs_ablation_combo}
\end{figure*}

\emph{Batch size.} Increasing batch size in our scheme consistently improves RM performance. Each batch of $B$ prefix--suffix pairs provides $B$ positives and $B(B-1)$ cross-pair negatives, so scaling $B$ increases ranking supervision per update nearly quadratically. Accordingly, larger batches yield higher peak gains over the initialized seed on RewardBench 2 (\cref{fig:bs_ablation_combo}a): ID improvements scale cleanly with $B$, while OOD gains plateau with little benefit beyond $B{=}16$. Larger $B$ also reaches higher accuracy with a more stable trajectory over the 11M-token budget (\cref{fig:bs_ablation_combo}b), with ID stability and final performance continuing to improve even as OOD curves largely saturate after $B{=}16$.

\ifdefined\arxivwrappedlayout
\begin{figure*}[!htbp]
\else
\begin{figure*}[t]
\fi
\centering

\begin{minipage}[t]{0.34\textwidth}\vspace{0pt}
  \centering
  \small
  \setlength{\tabcolsep}{2pt}
  \renewcommand{\arraystretch}{1.05}
\vskip 0.2in
  \resizebox{\linewidth}{!}{%
    \begin{tabular}{lrrrr}
      \toprule
      \textbf{Splitting} &
      \makecell{\textbf{ID}} &
      \makecell{\textbf{OOD}\\\textbf{Safety}} &
      \makecell{\textbf{OOD}\\\textbf{others}} &
      \makecell{\textbf{RewardBench 2}} \\
      \midrule
      \makecell[l]{preserve\\sentence} & +4.6 & +1.7 & +3.5 & +0.8 \\
      \rowcolor{rowgray}
      \makecell[l]{break\\sentence}    & \textbf{+16.1} & \textbf{+5.4} & \textbf{+7.4} & \textbf{+7.7} \\
      \bottomrule
    \end{tabular}%
  }

  \vspace{2pt}
  {\footnotesize (a) Peak RewardBench 2 subsets and average gains vs. data splitting design.}
\end{minipage}\hfill%
\begin{minipage}[t]{0.64\textwidth}\vspace{0pt}
  \centering
  \includegraphics[width=\linewidth]{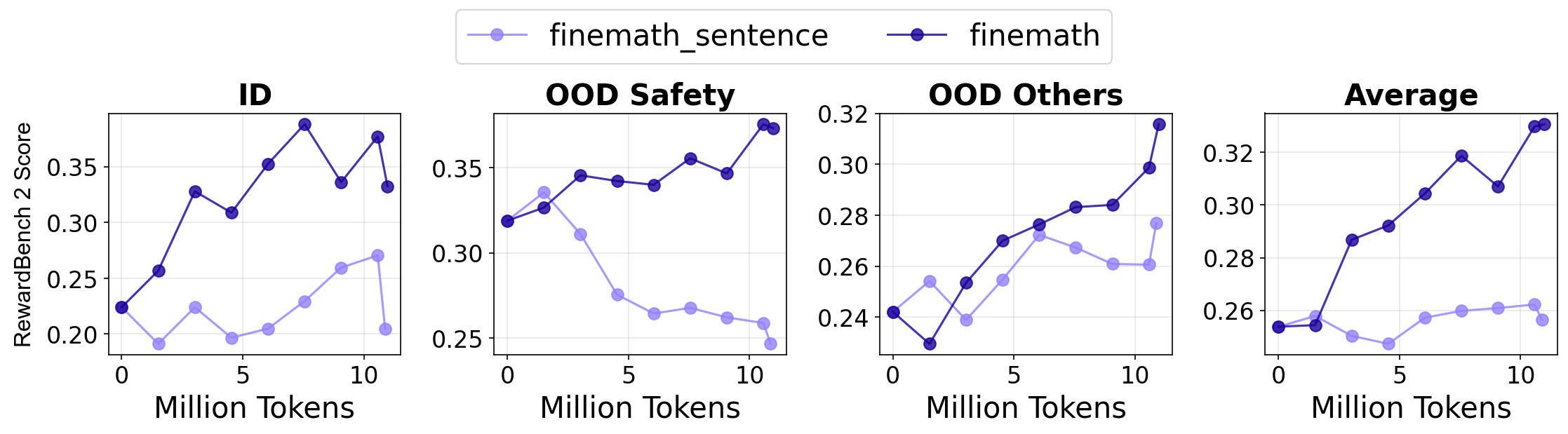}

  \vspace{2pt}
  {\footnotesize (b) RewardBench 2 learning curves across 11M token budget vs. data splitting design.}
\end{minipage}

\caption{Effect of data splitting design on peak gains and learning trajectory of RewardBench 2.}
\label{fig:splitting_ablation_combo}
\end{figure*}

\emph{Data splitting format.} Our online preference data depends on how we split raw web text into prefix--suffix pairs: either only at sentence boundaries (\emph{preserve sentence}) or potentially mid-sentence (\emph{break sentence}), which changes the difficulty of the induced in-batch negative samples. Allowing sentence breaks yields much larger gains over the initialized seed on RewardBench 2 and across all ID/OOD subsets (\cref{fig:splitting_ablation_combo}a,b). We attribute this to harder negative examples: in the \emph{preserve sentence} setting, packing prefix--suffix sequences while enforcing boundary constraints inevitably discards many candidate spans, leaving resulting batches with examples from disparate passages and making negatives less contextually similar and easier to reject. In contrast, when sentence breaks are allowed, many in-batch sequences remain contiguous chunks from the same underlying web text context, so cross-pair negatives tend to be more contextually confusable and require finer-grained semantic and contextual consistency to rank correctly. \Cref{sec:ex_sentence_splitting} provides more details on the \emph{preserve sentence} splitting algorithm and intuition of its failure.

\ifdefined\arxivwrappedlayout
\begin{figure*}[!htbp]
\else
\begin{figure*}[t]
\fi
\centering

\begin{minipage}[t]{0.36\textwidth}\vspace{0pt}
  \centering
  \small
  \setlength{\tabcolsep}{2pt}
  \renewcommand{\arraystretch}{1.05}

  \resizebox{\linewidth}{!}{%
    \begin{tabular}{lrrrrr}
      \toprule
      \textbf{Centering} &
      \makecell{\textbf{ID}} &
      \makecell{\textbf{OOD}\\\textbf{Safety}} &
      \makecell{\textbf{OOD}\\\textbf{others}} &
      \makecell{\textbf{RewardBench 2}} &
      \makecell{\textbf{BoN}} \\
      \midrule
      \makecell[l]{w/o\\centering} &
        \textbf{+16.9} & \textbf{+5.9} & \textbf{+8.6} & +6.4 & +2.8 \\
      \rowcolor{rowgray}
      \makecell[l]{w/\\centering} &
        +16.4 & +5.7 & +7.4 & \textbf{+7.7} & \textbf{+5.4} \\
      \bottomrule
    \end{tabular}%
  }

  \vspace{2pt}
  {\footnotesize (a) Peak RewardBench 2 subsets, average gains and BoN max achieved performance (MAP) gain vs. centering loss design.}
\end{minipage}\hfill%
\begin{minipage}[t]{0.62\textwidth}\vspace{0pt}
  \centering
  \includegraphics[width=\linewidth]{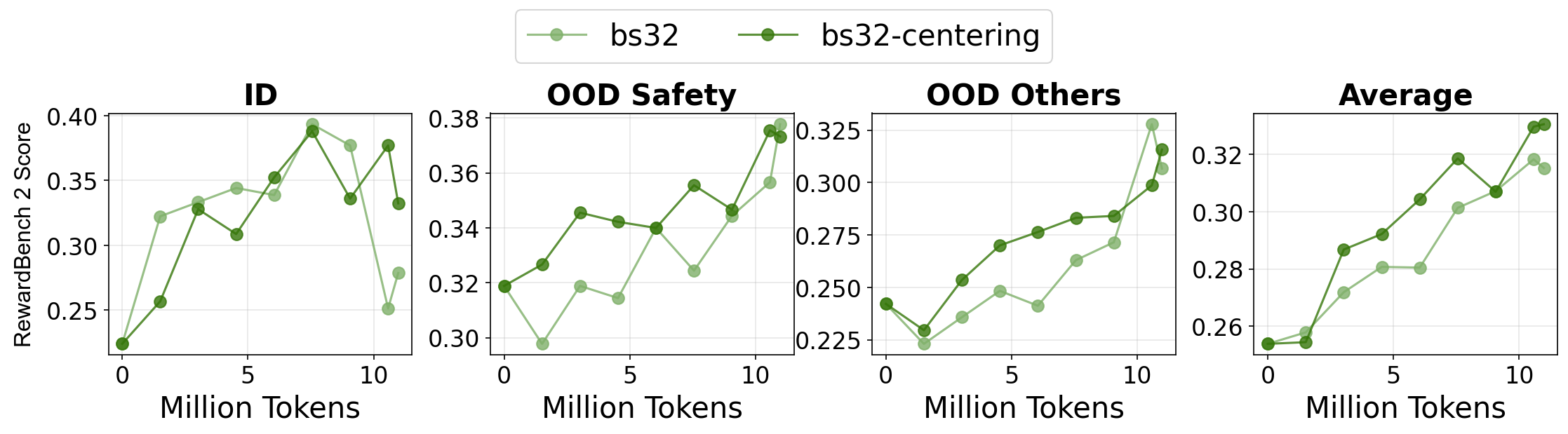}

  \vspace{2pt}
  {\footnotesize (b) RewardBench 2 learning curves across 11M token budget vs. centering loss.}
\end{minipage}

\caption{Effect of centering loss on RewardBench 2 peak gains, learning trajectory and BoN MAP.}
\label{fig:centering_ablation_combo}
\ifdefined\arxivwrappedlayout\else\vskip -0.2in\fi
\end{figure*}

\Needspace{17\baselineskip}
\begin{wrapfigure}{r}{0.35\textwidth}
\ifdefined\arxivwrappedlayout\vspace{-8pt}\fi
\centering
\includegraphics[width=\linewidth]{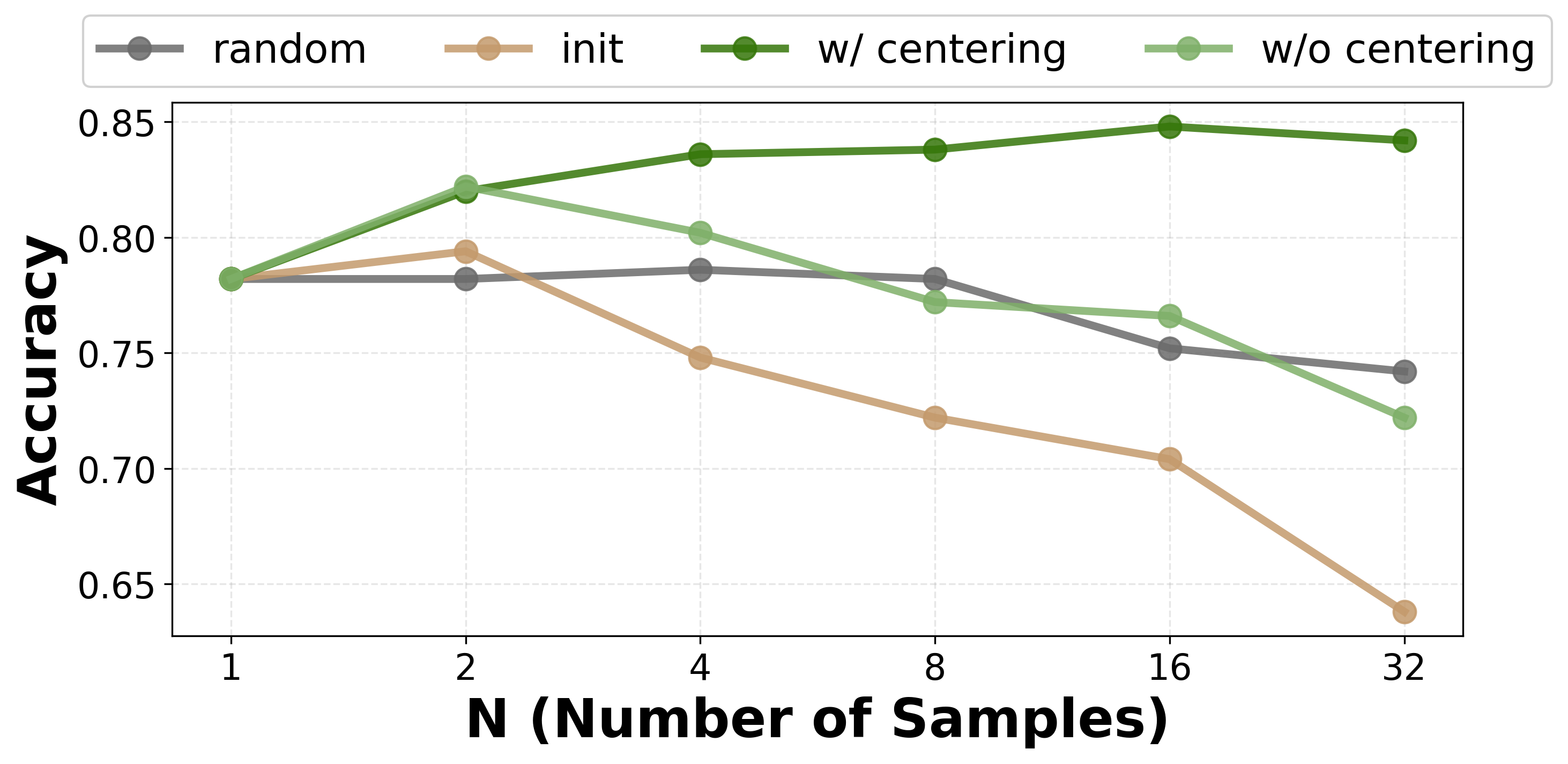}
\captionsetup{justification=raggedright,singlelinecheck=false}
\caption{BoN accuracy curve on \texttt{Llama-3.1-8B-Instruct} GSM8K rollouts for RMs trained with or without centering loss (\texttt{init}--initialized seed).}
\label{fig:bon_ablation}
\vspace{-0.1in}
\end{wrapfigure}

\noindent \emph{Centering loss.} We ablate the centering regularizer in our continuation-based Bradley--Terry RM objective by training with and without the quadratic penalty $\mathcal{L}_{\text{center}}$ (\cref{eq:center_loss}), which keeps chosen and in-batch rejected scores near zero and limits reward-scale drift on noisy web text. Training with centering yields a modest improvement in average RewardBench 2 performance, with peak gains over initialization of +7.7 versus +6.4 points without centering (\cref{fig:centering_ablation_combo}a,b). Centering provides a larger benefit for downstream response selection, measured by best-of-$N$ (BoN), which selects the highest-scoring response among $N$ actor-generated candidates (\cref{sec:bon}). On GSM8K \citep{cobbe2021trainingverifierssolvemath} with a \modelname{Llama-3.1-8B-Instruct} actor, centering substantially boosts the gain in Maximum Achieved Performance ($\Delta$MAP) over initialization (+5.4 vs.\ +2.8, \cref{fig:centering_ablation_combo}a). The benefit also grows with selection strength: the centered RM's accuracy increases monotonically with $N$ before saturating, while the uncentered RM peaks at small $N$ and eventually falls below the random baseline (\cref{fig:bon_ablation}).\par
\WFclear
\Needspace{5\baselineskip}
\emph{Continued-pretraining control.} To isolate the effect of the training objective, we continue pretraining \modelname{Llama-3.2-3B} with next-token prediction on the same 11M \textit{FineMath} tokens in the same order as NTP training. Since this model has no reward head, we compare two readouts trained on 512 auxiliary preference pairs from \textit{FineMath} blocks excluded from the earlier reward-model and continued-pretraining runs. Each pair uses a document prefix as the prompt, its original continuation as chosen, and a perturbed copy as rejected; evaluation uses RewardBench 2. The \emph{frozen-backbone probe} fixes the backbone and fits a linear scoring head to the final-layer hidden state at the last response token. We fit it with the Bradley--Terry objective to score original continuations above perturbed copies. \emph{Full reward-model conversion} jointly finetunes a newly initialized scalar reward head and all backbone parameters on the same pairs for three epochs with the same objective. We apply both readouts to the continued-pretrained and untouched base models (\cref{sec:ex_cpt}).\par

Continued pretraining on the same text does not improve reward-model performance over the untouched base model under either the frozen-backbone probe or full reward-model conversion (\cref{tab:cpt_control}). Replacing the NTP reward model's trained head with the same probe preserves its performance, indicating that the probe can recover reward signals learned by NTP. Together, these findings support the contribution of NTP preference learning beyond additional next-token training on the same corpus.\par

\Needspace{7\baselineskip}
\subsection[Generalization to Other Web Corpora]{Generalization to Other Web Corpora}
\label{sec:corpus_generalization}

\ifdefined\arxivwrappedlayout
\ifdefined\arxivwrappedlayout
\begin{wraptable}{r}{0.46\textwidth}
\else
\begin{wraptable}{R}{0.46\textwidth}
\fi
\ifdefined\arxivwrappedlayout\vspace{-8pt}\fi
\centering
\captionsetup{font=footnotesize,skip=3pt}
\caption{Peak average RewardBench 2 accuracy gains (percentage points) over the shared initialization for \modelname{Llama-3.2-3B} RMs trained on 11M tokens from each corpus.}
\label{tab:corpus_gain}
\scriptsize
\setlength{\tabcolsep}{3pt}
\renewcommand{\arraystretch}{1.12}
\begin{tabular*}{\linewidth}{@{\extracolsep{\fill}}lc@{}}
\toprule
\textbf{Training corpus} & \makecell{\textbf{RewardBench 2}\\\textbf{gain}} \\
\midrule
FineMath-4plus & $+7.7$ \\
MegaMath-Web-Pro-Max & $+8.3$ \\
FineWeb-Edu & $+7.6$ \\
\bottomrule
\end{tabular*}

\par
\end{wraptable}

\fi
To test whether NTP remains effective beyond \textit{FineMath-4plus}, we train on \textit{MegaMath-Web-Pro-Max} \citep{wang2025octothinkermidtrainingincentivizesreinforcement}, another mathematical web corpus, and \textit{FineWeb-Edu} \citep{penedo2024fineweb}, a general-domain educational web corpus. We use the same \modelname{Llama-3.2-3B} initialization and training settings as the \textit{FineMath-4plus} run, with 11M tokens per corpus.\par

\ifdefined\arxivwrappedlayout\else
\Needspace{13\baselineskip}

\fi
\noindent\Cref{tab:corpus_gain} reports peak average RewardBench 2 accuracy gains over initialization. \textit{FineMath-4plus}, \textit{MegaMath-Web-Pro-Max}, and \textit{FineWeb-Edu} yield gains of 7.7, 8.3, and 7.6 percentage points, respectively. \textit{MegaMath-Web-Pro-Max} gives the largest gain, exceeding \textit{FineMath-4plus} by 0.6 points, while general-domain \textit{FineWeb-Edu} comes within 0.1 points of \textit{FineMath-4plus}. These results show that NTP's effectiveness extends across web corpora and that substantial reward-model improvements can also be obtained from general-domain educational text.\par
\WFclear

\subsection{Best-of-N Accuracy and Scaling}
\label{sec:bon}

Motivated by ID and OOD RewardBench and RewardBench 2 gains, we evaluate our web-trained RMs using BoN selection. We sample $N \in \{1,2,4,8,16,32\}$ responses per prompt, select the highest-scoring one, and report Maximum Achieved Performance (MAP), the best task accuracy over $N$. We evaluate three actors, \modelname{Llama-3.2-1B-Instruct}, \modelname{Llama-3.2-3B-Instruct}, and \modelname{Llama-3.1-8B-Instruct}, on two in-domain mathematical reasoning tasks, MATH500 and GSM8K \citep{lightman2023letsverifystepstep,cobbe2021trainingverifierssolvemath}, and three out-of-domain tasks: HumanEval+ for code generation, ToxiGen for safety, and IFEval for instruction following \citep{chen2021evaluatinglargelanguagemodels,liu2023evalplus,hartvigsen2022toxigenlargescalemachinegenerateddataset,zhou2023instructionfollowingevaluationlargelanguage}. We use two RMs trained with our online continuation-based method (\cref{sec:algorithm}): \modelname{FineMath-RM-Llama-3.2-3B}, initialized from \modelname{Llama-3.2-3B}, and \modelname{FineMath-RM-Qwen-2.5-7B}, initialized from \modelname{Qwen2.5-7B-Instruct}. We compare against three off-the-shelf supervised RMs: \modelname{Skywork-Reward-V2-Llama-3.1-8B}, \modelname{Skywork-Reward-V2-Llama-3.2-3B} \citep{liu2025skyworkrewardv2scalingpreferencedata}, and PKU's \modelname{beaver-7b-unified-reward}, a safety-oriented RM \citep{dai2024saferlhf,ji2025pkusaferlhfmultilevelsafetyalignment}, as well as random selection. When measuring gains over initialization, we also score the same candidate sets with each RM's untrained initialization and report $\Delta$MAP. Evaluation details appear in \cref{sec:ex_bon}.

Across ID and OOD tasks, BoN gains become more consistent as actor size grows. Accuracy generally improves with $N$: gains are weakest for \modelname{FineMath-RM-Llama-3.2-3B} with the smallest \modelname{Llama-3.2-1B-Instruct} actor, but become clear for larger actors. \modelname{FineMath-RM-Qwen-2.5-7B} yields larger, often monotonic gains, especially with stronger actors. These results suggest that our RMs reliably select better responses across ID math reasoning and OOD code, safety, and instruction following. Actor-size scaling benefits \modelname{FineMath-RM-Llama-3.2-3B} more on ID than OOD tasks, while \modelname{FineMath-RM-Qwen-2.5-7B} shows similar trends across tasks (\cref{fig:summary_linear}).

\ifdefined\arxivwrappedlayout
\Needspace{32\baselineskip}
\else
\Needspace{31\baselineskip}
\fi
\begin{wrapfigure}{r}{0.52\textwidth}
\ifdefined\arxivwrappedlayout\vspace{-3pt}\fi
\centering
\includegraphics[width=\linewidth]{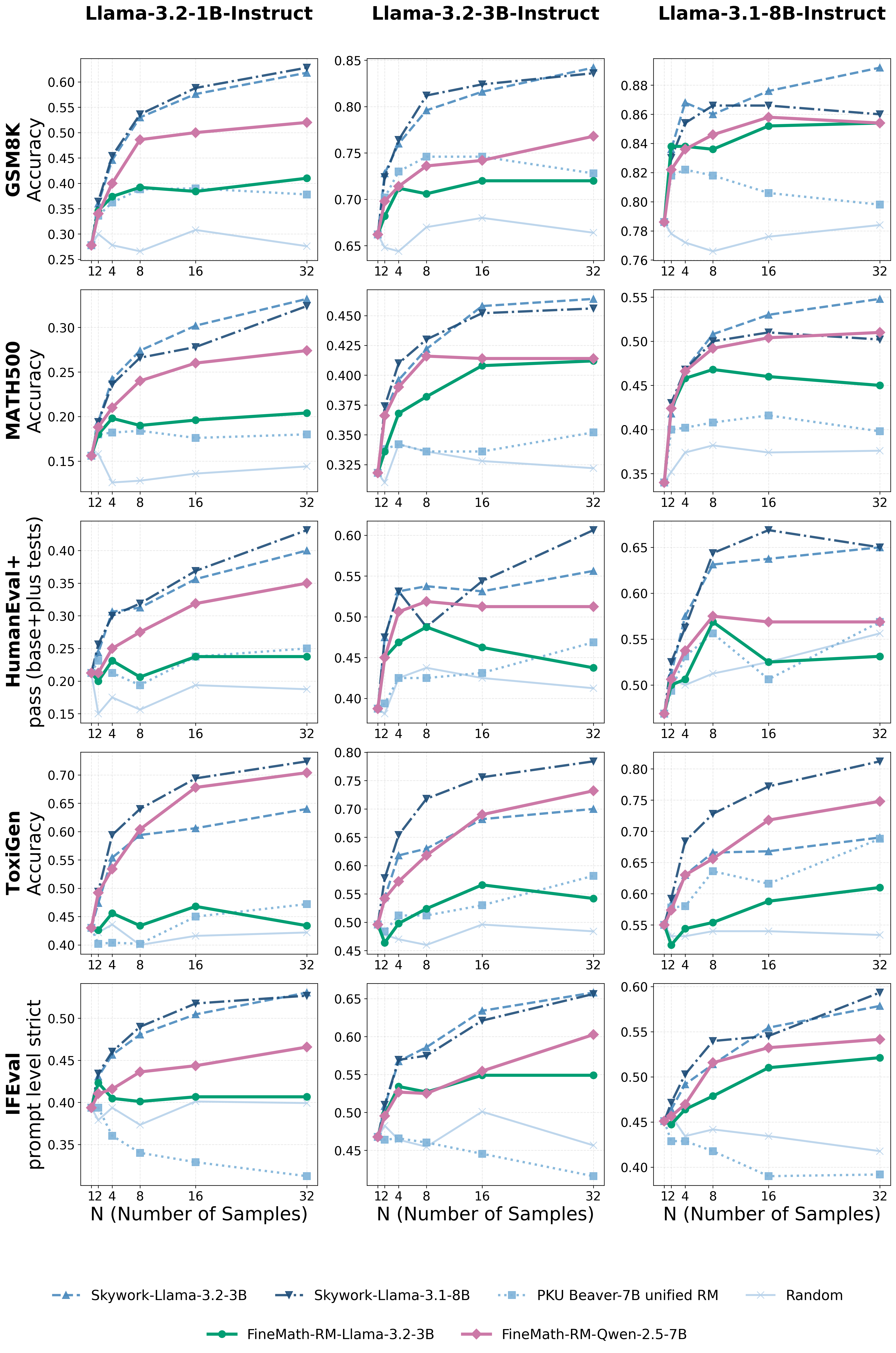}
\captionsetup{font=footnotesize,skip=3pt}
\caption{BoN scaling curves of RMs across two ID math tasks (GSM8K, MATH500) and three OOD tasks (HumanEval+, Toxigen, IFEval).}
\label{fig:summary_linear}
\end{wrapfigure}

Supervised RMs vary substantially in BoN performance. Across tasks and actors, the Beaver RM trails the stronger Skywork RM by 4 to 25 MAP points, so curated preference training alone does not determine selection quality. Our RMs become more competitive with the strongest supervised baselines as actor size grows. Moreover, \modelname{FineMath-RM-Qwen-2.5-7B} exceeds \modelname{Skywork-Reward-V2-Llama-3.2-3B} on OOD ToxiGen at every actor size and matches \modelname{Skywork-Reward-V2-Llama-3.1-8B} on ID MATH500 with the 8B actor (\cref{fig:summary_linear}). Compared with the Beaver RM, \modelname{FineMath-RM-Qwen-2.5-7B} achieves higher MAP on every task and actor size, while \modelname{FineMath-RM-Llama-3.2-3B} matches or exceeds it on GSM8K, MATH500, HumanEval+, and IFEval in all but one setting, with lower MAP mainly on ToxiGen. On IFEval, the Beaver RM performs worse than random selection as $N$ grows, whereas both of our RMs improve over the actor. The strongest curated-data RMs remain ahead overall, but our web-trained RMs sit inside the range spanned by supervised RMs of similar size without any human or LLM labels. These results indicate that raw math web text yields an effective selection signal that strengthens with RM and actor size.

\WFclear
\subsection{Actor Training}
\label{sec:actor_training}
\ifdefined\arxivwrappedlayout
\Needspace{14\baselineskip}
\else
\Needspace{22\baselineskip}
\fi
\begin{wraptable}{r}{0.56\textwidth}
\ifdefined\arxivwrappedlayout\vspace{-8pt}\else\vspace{-4pt}\fi
\centering
\setlength{\tabcolsep}{2pt}
\renewcommand{\arraystretch}{1.15}
\caption{Peak mean@1 test accuracy within five GRPO epochs. Both actors use instruction-tuned initializations. \textbf{Bold} and \underline{underlined} values denote the best and second-best results, respectively.}
\label{tab:rm_actor_tasks}

\scriptsize
\begin{tabularx}{\linewidth}{@{}>{\raggedright\arraybackslash}X S[table-format=1.3] S[table-format=1.3] S[table-format=1.3] S[table-format=1.3]@{}}
\toprule
\multirow{2}{*}{} & \multicolumn{2}{c}{\makecell{\textbf{Actor:}\\\textbf{Llama-3.2-}\\\textbf{3B-Instruct}}} & \multicolumn{2}{c}{\makecell{\textbf{Actor:}\\\textbf{Llama-3.1-}\\\textbf{8B-Instruct}}} \\
\cmidrule(lr){2-3} \cmidrule(lr){4-5}
& {MATH} & {GSM8K} & {MATH} & {GSM8K} \\
\midrule
Before training & 0.268 & 0.789 & 0.423 & 0.876 \\
\midrule
\modelname{Skywork-Reward-V2-Llama-3.1-8B} & 0.419 & \textbf{0.833} & \underline{0.447} & 0.882 \\
\modelname{Skywork-Reward-V2-Llama-3.2-3B} & \textbf{0.439} & 0.819 & \textbf{0.465} & \underline{0.884} \\
\modelname{beaver-7b-unified-reward} & 0.416 & 0.803 & 0.430 & 0.881 \\
\midrule
\textbf{\modelname{FineMath-RM-Qwen-2.5-7B}} & \underline{0.420} & \underline{0.823} & 0.437 & \textbf{0.886} \\
\addlinespace[2pt]
\modelname{FineMath-RM-Qwen-2.5-7B-Init} & 0.399 & 0.812 & 0.428 & 0.879 \\
\bottomrule
\end{tabularx}

\vspace{-4pt}
\end{wraptable}

Because \modelname{FineMath-RM-Qwen-2.5-7B} reliably selects stronger mathematical responses in BoN, we also test it as a reward for policy training. We train \modelname{Llama-3.2-3B-Instruct} and \modelname{Llama-3.1-8B-Instruct} with GRPO \citep{shao2024deepseekmathpushinglimitsmathematical} on MATH and GSM8K for five epochs, comparing against the two Skywork RMs, \modelname{beaver-7b-unified-reward} \citep{dai2024saferlhf,ji2025pkusaferlhfmultilevelsafetyalignment}, and our RM's untrained initialization. We report peak mean@1 test accuracy achieved within the same compute budget(\cref{sec:grpo_objective,sec:ex_actor_results}).

Our RM improves both actors on both tasks, achieves the best or second-best peak accuracy in three of the four settings, and gives the best GSM8K result for the 8B actor (\cref{tab:rm_actor_tasks}). It exceeds the Beaver RM by 0.4--2.0 percentage points across all four settings, showing that NTP training on raw web text provides an effective policy-training reward compared with this supervised baseline. It also outperforms its untrained initialization in every setting, supporting a consistent benefit from reward learning.

\WFclear
\ifdefined\arxivwrappedlayout\Needspace{5\baselineskip}\else\Needspace{9\baselineskip}\fi
\subsection{Mid-Training for Reward Modeling}
\label{subsec:midtraining}

Although NTP is competitive with supervised reward training under a matched token budget (\cref{tab:curated_comparison}), our models still in many cases underperform the strongest reward models trained on large curated preference datasets. We therefore also study NTP as a complement to supervised training, rather than a head-to-head comparison, and ask whether an initialization learned from unlabeled web text can reduce the need for curated preference labels.\par

\ifdefined\arxivwrappedlayout
\ifdefined\arxivwrappedlayout
  \def\rewardmixwidth{0.46\textwidth}
\else
  \def\rewardmixwidth{0.46\textwidth}
\fi
\begin{wraptable}{r}{\rewardmixwidth}
\vspace{-6pt}
\centering
\caption{RewardBench 2 gains from training on curated preference dataset \texttt{RLHFMix1} under different initializations.}
\scriptsize
\setlength{\tabcolsep}{2pt}
\renewcommand{\arraystretch}{1.05}

\resizebox{\linewidth}{!}{%
  \begin{tabular}{lrrrr}
    \toprule
    \textbf{Checkpoint} &
    \makecell{\textbf{ID}} &
    \makecell{\textbf{OOD}\\\textbf{Safety}} &
    \makecell{\textbf{OOD}\\\textbf{others}} &
    \makecell{\textbf{RewardBench 2}} \\
    \midrule
    \makecell[l]{Llama-3.2-3B} &
      +14.5 & +22.3 & +1.6 & +8.3 \\
    \makecell[l]{Llama-3.2-3B-Instruct} &
      +19.4 & +22.6 & \textbf{+7.8} & \textbf{+13.1} \\
      \makecell[l]{Finemath-RM-Llama-3.2-3B} &
      \textbf{+21.4} & \textbf{+24.4} & +2.0 & +10.4 \\
    \bottomrule
  \end{tabular}%
}

\vspace{2pt}

\label{tab:rewardmix1}
\vspace{-8pt}
\end{wraptable}

\fi
We test whether the 11M-token FineMath RM checkpoint provides a useful initialization for subsequent curated preference learning. We compare three \modelname{Llama-3.2-3B} starting points: the base model, our NTP-trained RM, and the instruction-tuned model. Each receives the same subsequent training on \texttt{RLHFMix1}, a uniformly sampled mixture of UltraFeedback, HH-RLHF, SHP, and Summarize-from-Feedback \citep{dong2024rlhf,cui2024ultrafeedbackboostinglanguagemodels,bai2022training,pmlr-v162-ethayarajh22a,stiennon2022learningsummarizehumanfeedback}. We evaluate all resulting RMs on RewardBench 2, reporting overall, ID-Math, OOD-Safety, and OOD-Others accuracy using the split defined in \cref{sec:preference_alignment}; training details are in \cref{sec:ex_rlhfmix1}.

\ifdefined\arxivwrappedlayout\else
\Needspace{12\baselineskip}
\ifdefined\arxivwrappedlayout
  \def\rewardmixwidth{0.46\textwidth}
\else
  \def\rewardmixwidth{0.46\textwidth}
\fi
\begin{wraptable}{r}{\rewardmixwidth}
\vspace{-6pt}
\centering
\caption{RewardBench 2 gains from training on curated preference dataset \texttt{RLHFMix1} under different initializations.}
\scriptsize
\setlength{\tabcolsep}{2pt}
\renewcommand{\arraystretch}{1.05}

\resizebox{\linewidth}{!}{%
  \begin{tabular}{lrrrr}
    \toprule
    \textbf{Checkpoint} &
    \makecell{\textbf{ID}} &
    \makecell{\textbf{OOD}\\\textbf{Safety}} &
    \makecell{\textbf{OOD}\\\textbf{others}} &
    \makecell{\textbf{RewardBench 2}} \\
    \midrule
    \makecell[l]{Llama-3.2-3B} &
      +14.5 & +22.3 & +1.6 & +8.3 \\
    \makecell[l]{Llama-3.2-3B-Instruct} &
      +19.4 & +22.6 & \textbf{+7.8} & \textbf{+13.1} \\
      \makecell[l]{Finemath-RM-Llama-3.2-3B} &
      \textbf{+21.4} & \textbf{+24.4} & +2.0 & +10.4 \\
    \bottomrule
  \end{tabular}%
}

\vspace{2pt}

\label{tab:rewardmix1}
\vspace{-8pt}
\end{wraptable}

\fi

As expected, instruct initialization outperforms base initialization, consistent with standard practice of starting RM training from an SFT model \citep{ouyang2022training,bai2022training}. Crucially, our RM checkpoint recovers much of this benefit: it achieves +10.4 average gain on RewardBench 2, matching or exceeding instruct on key subcategories---OOD-Safety (+24.4 vs. +22.6) and ID (+21.4 vs. +19.4)---while substantially outpacing the base model on ID (+14.5) (\cref{tab:rewardmix1,fig:rlhfmix1_curve}). This indicates that our unsupervised RM training provides a particularly strong foundation for learning in-distribution preferences from curated data. Gains on OOD Others are modest, likely due to forgetting induced by distribution shift from web text to curated data; in OOD Safety, a similar initial drop is recovered and ultimately surpassed (\cref{fig:rlhfmix1_curve}).
\par
\WFclear
To assess whether NTP initialization reduces the need for curated supervision, we vary the amount of \texttt{RLHFMix1} data (\cref{sec:ex_label_efficiency}). NTP initialization outperforms base initialization at every evaluated budget. With 50\% of the labeled data (approximately 122K pairs), it reaches an overall RewardBench 2 score of approximately 44, matching its own full-data result and exceeding the end performance of base initialization with all 244K pairs.\par

\section{Related Work}
\label{sec:related}

\textbf{Sources of reward supervision.}
Conventional reward models learn to rank candidate responses from human comparisons \citep{christiano2017deep,stiennon2020learning,ouyang2022training,bai2022training}.
The human effort required to collect these comparisons motivates replacing some human evaluations with LLM judgments \citep{zheng2023judging,dubois2023alpacafarm}.
Constitutional AI \citep{bai2022constitutional}, UltraFeedback \citep{cui2024ultrafeedbackboostinglanguagemodels}, and Self-Rewarding Language Models \citep{yuan2025selfrewardinglanguagemodels} use such judgments as feedback, while Semi-Supervised Reward Modeling (SSRM) \citep{he-etal-2024-semi-supervised} extends a labeled seed set with high-confidence comparisons pseudo-labeled by a reward model.
Although these methods reduce human labeling, generating candidate responses and running LLM judges still incur monetary costs, as illustrated by our estimates in \cref{sec:ex_cost}.
LLM judgments also remain susceptible to position bias \citep{shi2025judging} and non-transitive preferences \citep{xu2025investigating}.
Other methods derive supervision from model behavior without explicit preference judgments.
POLAR \citep{dou2025polar}, for example, pretrains on model-generated responses, using whether they come from the same policy to construct training labels.
Unsupervised Process Reward Models (uPRM) \citep{gadetsky2026uprm} derive supervision for reasoning-step evaluation from a pretrained LLM's token probabilities.
Both approaches therefore retain a dependence on model-generated responses or model-based assessments to supply their training signal.
NTP instead constructs comparisons directly from existing web text, treating each prefix's original continuation as preferred over mismatched continuations.
The text itself thus supplies both responses and preference labels, avoiding the generation and judging costs of preference-data collection.
Concurrent work applies a similar idea to multimodal reward modeling by deriving preference pairs from web captions and their perturbations \citep{li2026unsupervisedmultimodal}.
Raw text has also served as a source of reinforcement signal for training policies rather than reward models. Reinforcement Pre-Training (RPT) \citep{dong2025reinforcementpretraining} and RLP \citep{hatamizadeh2025rlp} reward a generated chain of thought either for predicting the observed next token or for the information gain it provides about upcoming tokens, and use this signal to train the generating model itself. NTP instead uses continuation structure to learn a standalone scalar reward model that scores arbitrary responses.\par

\textbf{From text discrimination to quality scoring.}
NTP builds on discriminative pretraining, which extracts supervision by contrasting observed data with mismatched or corrupted alternatives.
Quick Thoughts \citep{logeswaran2018quickthoughts}, BERT's next-sentence prediction \citep{devlin2019bert}, and Contrastive Predictive Coding (CPC) \citep{oord2018cpc} use context matching or contrastive prediction to learn representations, while ELECTRA \citep{clark2020electrapretrainingtextencoders} learns by detecting generator-replaced tokens.
Where these tasks serve representation learning, NTP formulates continuation matching as Bradley--Terry preference learning and retains the response-level scalar scorer as a reward model.
Closer to NTP's scoring objective, BLEURT \citep{sellam-etal-2020-bleurt} pretrains a reference-based metric on perturbed text with automatic targets before fine-tuning on human ratings, while DEAM \citep{ghazarian-etal-2022-deam} learns dialogue-coherence scores from semantically manipulated examples.
NTP instead forms its preference pairs by matching and mismatching existing prefixes and continuations, requiring neither text perturbation nor semantic parsing.
The question is then whether learning this continuation relationship produces rewards that transfer to broader preference judgments and supervised reward training.
Such transfer also needs to hold when the scores guide model behavior, since proxy rewards can be exploited under optimization \citep{gao2022scaling,skalse2022defining}.
We therefore evaluate the learned rewards through best-of-$N$ selection and policy optimization as well as preference benchmarks (\cref{sec:exp}).\par

\Needspace{5\baselineskip}
\section{Concluding Remarks}
\label{sec:conclusions}

We introduced next-token preference (NTP), an unsupervised reward-modeling method that learns scalar rewards from web-text continuations without candidate generation or human or model-generated preference judgments.
Although this supervision comes from continuation consistency, the learned rewards transfer to broader judgments of response quality, including reasoning and out-of-domain safety after training on mathematical text.
Gains grow with training data and extend across model families and sizes.
Across mathematical and general-domain corpora, 11M tokens per corpus yield peak RewardBench 2 gains of 7.6--8.3 percentage points over initialization.
This transfer also has practical value. In best-of-$N$ selection and mathematical policy optimization, our reward models fall within the range spanned by supervised reward models of similar size, despite using no human or model-generated preference labels.\par

NTP also complements supervised reward learning by improving how effectively models use curated preference data.
With only half of the labeled \texttt{RLHFMix1} data, NTP initialization surpasses base initialization trained on the full dataset.
This complementary benefit suggests that NTP can serve as a form of reward-model pretraining, learning a useful reward prior from raw text before refinement with supervised preferences.
Scaling NTP training and combining it with targeted supervision are promising directions for further reducing dependence on expensive preference data.\par

\section*{Acknowledgments}
SK acknowledges the support from the National Science Foundation Grant under award IIS 2229881; JF, KB, SK and HZ acknowledge the Chan Zuckerberg Initiative Foundation for establishing the Kempner Institute for the Study of Natural and Artificial Intelligence. YL acknowledges Anvil AI and GPU allocations through allocation CIS230253 from the Advanced Cyberinfrastructure Coordination Ecosystem: Services \& Support (ACCESS) program, which is supported by U.S. National Science Foundation grants \#2138259, \#2138286, \#2138307, \#2137603, and \#2138296. JF is currently at Scale AI, but this work was conducted while at Harvard University.

{\small
\bibliographystyle{plainnat}
\bibliography{ref}
}

\clearpage
\appendix
\addcontentsline{toc}{section}{Appendix}
\renewcommand\thepart{}
\renewcommand\partname{}
\part{\texorpdfstring{\Large{\centerline{Appendices}}}{Appendices}}
\parttoc
\newpage
\let\arxivNeedspace\Needspace
\renewcommand{\Needspace}[1]{\arxivNeedspace{5\baselineskip}}
\let\arxivFloatBarrier\FloatBarrier
\renewcommand{\FloatBarrier}{}
\begingroup
\renewcommand{\topfraction}{0.9}
\renewcommand{\bottomfraction}{0.85}
\renewcommand{\textfraction}{0.08}
\ifdefined\arxivwrappedlayout
\renewcommand{\floatpagefraction}{0.9}
\else
\renewcommand{\floatpagefraction}{0.7}
\fi
\setlength{\textfloatsep}{12pt plus 2pt minus 2pt}
\setlength{\floatsep}{10pt plus 2pt minus 2pt}
\setlength{\intextsep}{10pt plus 2pt minus 2pt}
\captionsetup{font=small,skip=5pt}
\captionsetup[table]{position=top}
\makeatletter
\setlength{\@fptop}{0pt}
\setlength{\@fpsep}{12pt plus 2pt minus 2pt}
\setlength{\@fpbot}{0pt plus 1fil}
\makeatother

\section{Limitations and Broader Impacts}
\label{sec:limitations_impacts}
Although NTP is competitive with supervised reward training under a matched token budget, our models still underperform the strongest reward models trained on large curated preference datasets in many settings (\cref{tab:curated_comparison,sec:bon}). We therefore do not claim that NTP replaces supervised reward training. However, our mid-training experiments show that NTP can improve the label efficiency of supervised training by providing an effective initialization learned from unlabeled web text. With only 50\% of the labeled \texttt{RLHFMix1} data, NTP initialization surpasses base initialization trained on the full dataset and matches its own full-data performance on RewardBench 2 (\cref{subsec:midtraining,sec:ex_label_efficiency}). This result supports using unsupervised reward learning to substantially reduce the amount of annotated data needed for subsequent training while retaining the benefits of high-quality preference supervision.

The broader impact of this work is to make reward-model training more accessible and scalable by reducing the cost of constructing preference data (\cref{sec:ex_cost}). NTP constructs comparisons directly from existing web text without candidate generation or human or model-generated preference judgments. Together with the improved label efficiency of subsequent supervised training, this offers a way for limited annotation budgets to support stronger reward models. However, continuation consistency is an imperfect proxy for response quality: web text can contain factual errors, harmful content, and social biases that the learned reward function may reproduce. Better benchmark performance does not guarantee safe behavior under downstream optimization. Application-specific evaluation, checks for biased preferences and reward exploitation, and targeted human feedback remain important when deploying these models.

\FloatBarrier
\Needspace{5\baselineskip}
\section{Reproducibility: Dataset and Code}
\label{sec:code}

The study uses publicly available datasets listed in \cref{sec:licenses}. Our code is available at \url{https://github.com/jingxuanf0214/rm-scaling}. The algorithm, evaluation protocols, and training details are documented in \cref{sec:methods,sec:ex_data,sec:ex_bon,sec:ex_actor,sec:ex_rlhfmix1}.

\FloatBarrier
\Needspace{5\baselineskip}
\section{LLM Usage}
\label{sec:llm_usage}

LLMs were used in a supporting role during the preparation of this paper.
Specifically, we used LLM assistance for: (1) checking grammar errors and editing text for clarity and readability; and (2) helping write code to visualize experimental results (e.g., plotting scripts for figures).
LLM assistance was not used for any part of the core research methodology and algorithm design.

\FloatBarrier
\Needspace{5\baselineskip}
\section{Extended Related Works}
\label{sec:ex_rw}
\FloatBarrier
\Needspace{5\baselineskip}
\subsection{Cost Estimate of Current RLHF Pipeline}
\label{sec:ex_cost}

RLHF relies critically on the availability of high-quality preference datasets, but collecting those datasets is widely viewed as one of the dominant bottlenecks in the RLHF workflow---spanning prompt curation, candidate generation, preference labeling, and iterative refinement. This is emphasized in \cite{dong2024rlhf}, which lays out end-to-end RLHF pipelines and highlights how feedback collection/labeling becomes a central practical constraint when scaling RLHF beyond small offline datasets. 

In this context, we explicitly quantify the cost of the full preference-data workflow focusing primarily on candidate generation (a prompt is used to generate a candidate pool) and preference annotation (a judge assigns which candidate is chosen/rejected) using representative datasets used by the \cite{dong2024rlhf} RLHF workflow. 

Because the underlying datasets may use a mixture of completion models and judge models, we report costs by instantiating token volumes under a single representative model class for each stage. Specifically, for both candidate generation and LLM judging, we assume a GPT-4–class model for demonstration (e.g., \texttt{gpt-4o} \citep{openai2024gpt4ocard}), not because it necessarily reflects the exact model used in data collection, but because it provides a consistent reference point for pricing across datasets. This assumption should be viewed as an average-cost proxy—rather than the most advanced or specialized reasoning model—while still being strong enough to produce high-quality candidate completions and reliable preference judgments.

To estimate candidate generation cost, we treat each preference pair as one prompt that yields two generated completions (preferred and rejected). Thus, each pair corresponds to a single generation call with prompt-side input (plus optional fixed overhead) and two completion outputs. Using the dataset length statistics, the generated output tokens per pair are computed as:
\begin{equation}
T_{\text{cand}} = \text{PrefLen} + \text{RejLen}.
\end{equation}

We then include an assumed constant input overhead $O_{\text{in}}^{\text{gen}}$ (e.g., system message + formatting constraints). For a dataset with $N$ preference pairs, the total token usage for candidate generation is:
\begin{equation}
T_{\text{in}}^{\text{gen}} = N \big(\text{PromptLen} + O_{\text{in}}^{\text{gen}}\big), \quad T_{\text{out}}^{\text{gen}} = N \cdot T_{\text{cand}}.
\end{equation}

Finally, given per-million token prices $P_{\text{in}}^{\text{gen}}$ and $P_{\text{out}}^{\text{gen}}$ for a GPT-4--class completion model, the total candidate generation cost is:
\begin{equation}
\label{eq:gen_cost}
\text{Cost}^{\text{gen}} = (T_{\text{in}}^{\text{gen}} / 10^6) P_{\text{in}}^{\text{gen}} + (T_{\text{out}}^{\text{gen}} / 10^6) P_{\text{out}}^{\text{gen}}.
\end{equation}

In \cref{tab:candidate_gen_cost_all}, we report candidate generation cost estimates based on the above calculations \cref{eq:gen_cost} for the representative preference dataset included in \cite{dong2024rlhf}. Costs are computed per dataset using the publicly listed \texttt{gpt-4o} text rates $P_{\text{in}} = \$2.50$ and $P_{\text{out}} = \$10.00$ per 1M tokens. We additionally report the aggregate total obtained by summing costs across datasets, reflecting the common RLHF practice of training on the union of these curated preference corpora.

\begin{table}[!htbp]
\centering
\caption{
Estimated candidate generation cost instantiated with a GPT-4--class completion model \texttt{gpt-4o}.
}
\label{tab:candidate_gen_cost_all}
\small
\setlength{\tabcolsep}{2pt}
\resizebox{\textwidth}{!}{
\begin{tabular}{lrrrrrrrrrr}
\toprule
Dataset & \#Pairs $N$ & \makecell{Prompt\\length} & \makecell{Preferred\\length} & \makecell{Rejected\\length} &
$T_{\text{cand}}$ &
\makecell{Overhead\\(in/out)} &
$T_{\text{in}}^{\text{gen}}$ &
$T_{\text{out}}^{\text{gen}}$ &
Cost (\$) &
\$/pair \\
\midrule
HH-RLHF &
115{,}396 &
160.4 & 82.2 & 73.6 &
155.8 &
$0$ / -- &
18.51M &
17.98M &
180.85 &
0.00157 \\
SHP &
93{,}301 &
186.2 & 173.6 & 88.8 &
262.4 &
$0$ / -- &
17.37M &
24.48M &
230.60 &
0.00247 \\
HelpSteer &
37{,}131 &
530.0 & 116.4 & 89.3 &
205.7 &
$0$ / -- &
19.68M &
7.64M &
100.46 &
0.00271 \\
PKU-SafeRLHF-30K &
26{,}874 &
21.5 & 70.4 & 74.6 &
145.0 &
$0$ / -- &
0.58M &
3.90M &
32.33 &
0.00120 \\
UltraFeedback &
340{,}025 &
161.5 & 279.5 & 211.1 &
490.6 &
$0$ / -- &
54.91M &
166.82M &
1444.36 &
0.00425 \\
UltraInteract &
161{,}927 &
507.4 & 396.6 & 416.7 &
813.3 &
$0$ / -- &
82.16M &
131.70M &
1217.89 &
0.00752 \\
CodeUltraFeedback &
50{,}156 &
172.8 & 427.6 & 400.6 &
828.2 &
$0$ / -- &
8.67M &
41.54M &
349.65 &
0.00697 \\
Argilla-Math &
2{,}418 &
36.5 & 276.5 & 265.3 &
541.8 &
$0$ / -- &
0.09M &
1.31M &
10.66 &
0.00441 \\
OpenOrca &
6{,}926 &
153.3 & 165.4 & 260.5 &
425.9 &
$0$ / -- &
1.06M &
2.95M &
25.72 &
0.00371 \\
Capybara &
14{,}811 &
634.5 & 348.4 & 401.9 &
750.3 &
$0$ / -- &
9.40M &
11.11M &
107.70 &
0.00727 \\
\midrule
\cite{dong2024rlhf} &
-- & -- & -- & -- & -- & -- & -- & -- &
3700.22 &
-- \\
\bottomrule
\end{tabular}
}

\end{table}

To estimate the LLM annotation cost (assuming single judge), we treat each preference pair as one judge call whose input is the concatenation of the prompt and the two candidate responses plus a fixed rubric/formatting overhead, and whose output is a short structured decision. The content tokens per pair are computed as:
\begin{equation}
    T_{\text{content}} = \text{PromptLen} + \text{PrefLen} + \text{RejLen}.
\end{equation}

We then include an assumed constant input overhead $O_{\text{in}}^{judge}$ tokens and output budget $O_{\text{out}}^{judge}$ tokens (e.g., JSON verdict + brief rationale). For a dataset with $N$ labeled pairs, total token usage is:
\begin{equation}
    T_{\text{in}}^{judge} = N (T_{\text{content}} + O_{\text{in}}^{judge}), \quad T_{\text{out}}^{judge} = N \cdot O_{\text{out}}^{judge}.
\end{equation}

Given per-million token prices $P_{\text{in}}^{judge}, P_{\text{out}}^{judge}$ for a GPT-4--class judge, the total cost is:
\begin{equation}
\label{eq:ann_cost}
    \text{Cost} = (T_{\text{in}}^{judge} / 10^6) P_{\text{in}}^{judge} + (T_{\text{out}}^{judge} / 10^6) P_{\text{out}}^{judge}.
\end{equation}

Similar to \cref{tab:candidate_gen_cost_all}, \cref{tab:llm_annotation_cost_all} reports candidate annotation cost estimates based on calculation of \cref{eq:ann_cost} for the representative preference dataset included in \cite{dong2024rlhf}. 

\begin{table}[!htbp]
\centering
\caption{
Estimated LLM annotation cost for preference labeling with a single GPT-4--class judge (\texttt{gpt-4o} pricing).
We use $O_{\text{in}}=40$ input overhead tokens and $O_{\text{out}}=40$ output tokens per pair.
}
\label{tab:llm_annotation_cost_all}
\small
\setlength{\tabcolsep}{2pt}
\resizebox{\textwidth}{!}{
\begin{tabular}{lrrrrrrrrrr}
\toprule
Dataset & \#Pairs $N$ & \makecell{Prompt\\length} & \makecell{Preferred\\length} & \makecell{Rejected\\length} &
$T_{\text{content}}$ &
\makecell{Overhead\\(in/out)} &
$T_{\text{in}}^{\text{judge}}$ &
$T_{\text{out}}^{\text{judge}}$ &
Cost (\$) &
\$/pair \\
\midrule
HH-RLHF &
115{,}396 &
160.4 & 82.2 & 73.6 &
316.2 &
40 / 40 &
41.10M &
4.62M &
148.92 &
0.00129 \\
SHP &
93{,}301 &
186.2 & 173.6 & 88.8 &
448.6 &
40 / 40 &
45.59M &
3.73M &
151.29 &
0.00162 \\
HelpSteer &
37{,}131 &
530.0 & 116.4 & 89.3 &
735.7 &
40 / 40 &
28.79M &
1.49M &
86.91 &
0.00234 \\
PKU-SafeRLHF-30K &
26{,}874 &
21.5 & 70.4 & 74.6 &
166.5 &
40 / 40 &
5.55M &
1.07M &
24.57 &
0.00091 \\
UltraFeedback &
340{,}025 &
161.5 & 279.5 & 211.1 &
652.1 &
40 / 40 &
235.71M &
13.60M &
725.83 &
0.00213 \\
UltraInteract &
161{,}927 &
507.4 & 396.6 & 416.7 &
1320.7 &
40 / 40 &
220.56M &
6.48M &
616.28 &
0.00381 \\
CodeUltraFeedback &
50{,}156 &
172.8 & 427.6 & 400.6 &
1001.0 &
40 / 40 &
52.20M &
2.01M &
150.52 &
0.00300 \\
Argilla-Math &
2{,}418 &
36.5 & 276.5 & 265.3 &
578.3 &
40 / 40 &
1.50M &
0.10M &
4.74 &
0.00196 \\
OpenOrca &
6{,}926 &
153.3 & 165.4 & 260.5 &
579.2 &
40 / 40 &
4.29M &
0.28M &
13.53 &
0.00195 \\
Capybara &
14{,}811 &
634.5 & 348.4 & 401.9 &
1384.8 &
40 / 40 &
21.51M &
0.59M &
56.53 &
0.00382 \\
\midrule
\cite{dong2024rlhf} &
-- & -- & -- & -- & -- & -- & -- & -- &
1979.10 &
-- \\
\bottomrule
\end{tabular}
}

\end{table}

\FloatBarrier
\Needspace{5\baselineskip}
\section{Extended Details on RM training}
\label{sec:ex_data}

\FloatBarrier
\Needspace{5\baselineskip}
\subsection{Dataset Parameters and Examples}
\label{sec:ex_dataset_params}

For \textit{FineMath}, we use the \textit{FineMath-4plus} subset released by HuggingFace \citep{allal2025smollm2smolgoesbig}. \textit{4plus} indicates that they are the selective higher quality entries in the dataset (as opposed to \textit{3plus}). A representative data example is shown in \ref{fig:finemath-example}.

For the general-domain experiment, we use \textit{FineWeb-Edu} \citep{penedo2024fineweb}, the educational subset of FineWeb selected by an educational-quality classifier. An example excerpt from the public release is shown in \cref{fig:fineweb-example}.

For the additional mathematical corpus, we use \textit{MegaMath-Web-Pro-Max} from OctoThinker \citep{wang2025octothinkermidtrainingincentivizesreinforcement}. This release filters MegaMath-Web using a fastText classifier trained on Llama-3.1-70B-Instruct quality annotations, then refines the selected documents with the same LLM. We construct our in-batch comparisons from these released texts without generating preference labels. An example entry from the public release is shown in \cref{fig:megamath-example}.

We use fixed-length chunks of $L=1536$ tokens, with a default $1{:}2$ split into a prompt prefix of $L_1=512$ tokens and a continuation of $L_2=1024$ tokens, in the notation of \cref{sec:methods}. These lengths are fixed across all main experiments. The split is determined by token position and need not coincide with a sentence boundary.

To assess sensitivity to split position, we keep the total length fixed at $L=1536$ and compare a shorter prefix ($1{:}2$), equal-length segments ($1{:}1$), and a longer prefix ($2{:}1$), corresponding to $(L_1,L_2)=(512,1024)$, $(768,768)$, and $(1024,512)$, respectively. All three configurations yield positive RewardBench 2 gains, averaging $+7.87 \pm 0.96$ (mean $\pm$ standard error), with no clear trend relating the split ratio to performance. We therefore retain the $1{:}2$ split as a simple default rather than treating the ratio as a critical tunable hyperparameter.

\begin{figure}[!htbp]
\centering
\setlength{\fboxsep}{10pt}
\colorbox{gray!15}{%
\begin{minipage}{0.92\linewidth}
\small
\textbf{\# Period Of Function}

\vspace{0.3em}
{\footnotesize by -2 views}

\vspace{0.4em}
Period of sinusoidal functions from equation. You can figure this out without looking at a graph by dividing with the frequency which in this case is 2.

\vspace{0.5em}
\textbf{\#\#\# Periodic functions are used throughout science to describe oscillations, waves and other phenomena that exhibit periodicity.}

\vspace{0.4em}
This is why this function family is also called the periodic function family. Therefore in the case of the basic cosine function $f(x)$, when this occurs we call the horizontal shift the period of the function. If a function has a repeating pattern like sine or cosine it is called a periodic function. Amplitude $a$: let $b$ be a real number. Horizontal stretch is measured for sinusoidal functions as their periods.

\vspace{0.4em}
For basic sine and cosine functions the period is $2\pi$. Any function that is not periodic is called aperiodic. The period is defined as the length of one wave of the function. The period is the length on the $x$-axis in one cycle. The period of a sinusoid is the length of a complete cycle.

\vspace{0.4em}
More formally, we say that this type of function has a positive constant $k$ where any input $x$:
\[
f(x + k) = f(x)
\]
The period is defined as the length of a function's cycle. According to the periodic function definition, the period of a function is represented as $f(x) = f(x + p)$, where $p$ is the real number equal to the period of the given function $f(x)$.

\vspace{0.4em}
Where $f(x + P) = f(x)$ for all values of $x$. The period of a periodic function is the interval of $x$-values on which the cycle of the graph that is repeated in both directions lies. In this case, one full wave is 180 degrees or $\pi$ radians. So the period of $\sin$ or $\cos$ is $2\pi$.

\vspace{0.4em}
The period of the function can be calculated using:
\[
P = \frac{2\pi}{|b|}
\]
In the problems below we will use the formula for the period $P$ of trigonometric functions of the form $y = a\sin(bx + c) + d$ or $y = a\cos(bx + c) + d$.

\vspace{0.4em}
The amplitude of $y = a\sin x$ and $y = a\cos x$ represents half the distance between the maximum and minimum values of the function. The period is the length of the smallest interval that contains exactly one copy of the repeating pattern. A periodic function repeats its values at set intervals called periods.
\end{minipage}%
}

\caption{\textit{FineMath-4plus} dataset example.}
\label{fig:finemath-example}
\end{figure}

\begin{figure}[!htbp]
\centering
\setlength{\fboxsep}{10pt}
\colorbox{gray!15}{%
\begin{minipage}{0.92\linewidth}
\small
\textbf{Hold the salt: UCLA engineers develop revolutionary new desalination membrane}

\vspace{0.5em}
\textbf{Process uses atmospheric pressure plasma to create filtering 'brush layer'}

\vspace{0.4em}
Desalination can become more economical and used as a viable alternate water resource.

\vspace{0.3em}
{\footnotesize By Wileen Wong Kromhout\\
Originally published in UCLA Newsroom}

\vspace{0.4em}
Researchers from the UCLA Henry Samueli School of Engineering and Applied Science have unveiled a new class of reverse-osmosis membranes for desalination that resist the clogging which typically occurs when seawater, brackish water and waste water are purified.

\vspace{0.4em}
The highly permeable, surface-structured membrane can easily be incorporated into today's commercial production system, the researchers say, and could help to significantly reduce desalination operating costs. Their findings appear in the current issue of the Journal of Materials Chemistry.

\vspace{0.4em}
Reverse-osmosis (RO) desalination uses high pressure to force polluted water through the pores of a membrane. While water molecules pass through the pores, mineral salt ions, bacteria and other impurities cannot. Over time, these particles build up on the membrane's surface, leading to clogging and membrane damage. This scaling and fouling places higher energy demands on the pumping system and necessitates costly cleanup and membrane replacement.

\vspace{0.4em}
\ldots
\end{minipage}%
}

\caption{\textit{FineWeb-Edu} dataset example (excerpt from \href{http://www.environment.ucla.edu/water/news/article.asp?parentid=6178}{UCLA Newsroom}).}
\label{fig:fineweb-example}
\end{figure}

\begin{figure}[!htbp]
\centering
\setlength{\fboxsep}{10pt}
\colorbox{gray!15}{%
\begin{minipage}{0.92\linewidth}
\small
\textbf{Scalars and Vectors}

Mathematics is the language of physics, allowing us to quantitatively describe the world. In mechanics, we use two types of quantities: scalars and vectors.

\vspace{0.4em}
\textbf{Goals:}
\begin{itemize}[leftmargin=*,nosep,topsep=0.2em]
    \item Learn the definitions of scalars and vectors
    \item See examples of vector and scalar quantities
    \item Translate vectors to and from their components
    \item Understand how to add, subtract, and multiply vectors
\end{itemize}

\vspace{0.4em}
\textbf{Definitions:}
\begin{itemize}[leftmargin=*,nosep,topsep=0.2em]
    \item Scalar: A scalar describes a one-dimensional quantity, requiring only one number to completely describe it. Examples include temperature, time, speed, mass, and location along a line (1D).
    \item Vector: A vector describes a multi-dimensional quantity, requiring more than one number to completely describe it. Vectors have two characteristics: magnitude and direction. Examples include location in a plane (2D), location in space (3D), velocity, acceleration, and force.
\end{itemize}

\vspace{0.4em}
\textbf{Vector Diagrams:}

Vector diagrams visualize physical systems. Although problems can be solved algebraically, diagrams help identify subtleties. The first step in solving physics problems is to draw a picture. Vector diagrams show the concept of the tip and tail of a vector, with direction indicated by an arrowhead. Two vectors with the same magnitude and direction are considered the same, allowing them to be moved around the coordinate system to simplify math.

\vspace{0.4em}
\textbf{Expressing Vectors Using Components:}

Vectors can be expressed in terms of magnitude and direction or components. The magnitude is the vector's length, and direction is given in terms of an angle. Vectors are often broken down into component vectors, which run parallel to the coordinate axes. A two-dimensional vector has two component vectors: one in the X direction and one in the Y direction.

\vspace{0.4em}
\textbf{Vector Addition:}

Vector addition involves moving the tail of the second vector to the tip of the first vector to get the resultant vector. Vectors can be added by expressing them as independent components and adding corresponding components.

\vspace{0.4em}
\textbf{Multiplication of a Vector by a Scalar:}

A vector can be multiplied by a scalar by multiplying each component by that number. The vector's direction remains the same, but its length changes. If $A = (1,2)$, then $3A = (3,6)$. Multiplying a vector by $-1$ reverses its direction.

\vspace{0.4em}
\textbf{Vector Subtraction:}

Vector subtraction works like vector addition, except the vector being subtracted is multiplied by $-1$. This is similar to subtracting numbers: $A - B = A + (-B)$. The order of vectors is important, as reversing them results in a vector of opposite direction.

\vspace{0.4em}
\textbf{Example 1:}

A hiker walks 53.1 degrees north of east for 2.5 km and then due east for 2.0 km. What is her total displacement from the starting point, measured along a straight line?

\vspace{0.4em}
\textbf{Example 2:}

A telephone pole support cable must be moved 2 meters closer to the pole. If the pole is 10 meters tall and the cable is currently fastened to the ground 8 meters from the pole, how much will the workers need to cut off from the cable when they move it?
\end{minipage}%
}

\caption{\textit{MegaMath-Web-Pro-Max} dataset example from the OctoThinker release. The entry contains LLM-refined text; its recorded source is an \href{http://webphysics.iupui.edu/JITTworkshop/152Basics/vectors/vectors.html}{IUPUI tutorial on scalars and vectors}.}
\label{fig:megamath-example}
\end{figure}

\FloatBarrier
\Needspace{5\baselineskip}
\subsection{Contamination Analysis}
\label{sec:ex_contamination}
\textit{FineMath-4plus}, used for our main experiments, was released as part of the SmolLM2 data pipeline \citep{allal2025smollm2smolgoesbig}, which reports 13-gram decontamination against several standard math and reasoning benchmarks, including GSM8K, MATH, MMLU, ARC-E, and ARC-C. Since GSM8K and MATH/MATH500 are among our main downstream math evaluations, this provides an existing contamination control for the core math benchmarks used in our study.

By contrast, \textit{FineWeb-Edu} \citep{penedo2024fineweb} inherits FineWeb's within-crawl MinHash deduplication based on 5-grams, but its release does not report benchmark-specific decontamination.

\textit{MegaMath-Web-Pro-Max} \citep{wang2025octothinkermidtrainingincentivizesreinforcement} is obtained by filtering and refining MegaMath-Web. The original MegaMath release \citep{zhou2025megamath} reports removing documents with exact 13-gram overlap against 12 math benchmarks, including GSM8K, MATH, MMLU, and AIME.

We further perform our own 13-gram contamination audit between the exact training subsets drawn from all three corpora and their corresponding evaluation benchmarks. For \textit{FineMath-4plus}, we audit RewardBench, IFEval, and ToxiGen, which are not covered by its source release; for \textit{FineWeb-Edu} and \textit{MegaMath-Web-Pro-Max}, we audit RewardBench 2. We find no 13-gram overlap in any of these corpus--benchmark comparisons. These results suggest that the reported gains are unlikely to be explained by direct benchmark leakage under this standard $n$-gram overlap criterion.

We will release the contamination-checking logs together with the code and training data processing scripts.

\FloatBarrier
\Needspace{5\baselineskip}
\subsection{Proxy-Judge Analysis of Training Signal Quality}
\label{sec:ex_proxy_judge}
\Cref{sec:methods} argues that in-batch cross-pairs serve as reasonable hard negatives on average. These comparisons target contextual compatibility: another continuation can occasionally fit the prefix, and the observed continuation is not guaranteed to be correct. However, the difficulty of a batch's implicit preference signal is not uniform: it depends on how contextually similar the in-batch negatives happen to be to each specific prefix. When negatives are drawn from passages that are topically or stylistically distant from the target prefix, the true continuation may be trivially distinguishable, providing little useful ranking supervision. Conversely, when negatives come from passages sharing close context—as encouraged by our non-sentence-aware splitting and larger batch sizes—the comparison becomes genuinely hard. This batch-level variation in signal difficulty can contribute to the noisy or non-monotonic gains observed during training, and provides an additional motivation for the design choices we ablate in \cref{sec:ablation}: larger batches increase the proportion of contextually similar in-batch negatives quadratically, and non-sentence-aware splitting retains more contiguous chunks from the same underlying passage, both of which raise the average difficulty of the training signal.

To quantify this effect, we conduct a diagnostic experiment using an external LLM as a proxy judge. We sample 100 training batches from the NTP training corpus, each with batch size 32. For each example in a batch, the prefix is paired with its true suffix as the positive continuation and with mismatched suffixes from other examples in the same batch as negative continuations. We then ask a \texttt{Llama-3.2-3B-Instruct} model to identify which candidate suffix is the coherent continuation of the given prefix. The proxy-judge accuracy for a batch is computed as the fraction of examples for which the model selects the true suffix over the in-batch negatives.

The resulting batch-level accuracies vary substantially, ranging from approximately 0.3 to 0.8 across the 100 sampled batches. This indicates that the implicit preference signal is heterogeneous: in some batches the true continuation is relatively easy to distinguish from mismatched suffixes, while in others the comparison is much more challenging. Critically, this heterogeneity also confirms that in-batch negatives are not uniformly trivial—many batches contain difficult comparisons where a small instruction-tuned model cannot reliably identify the true continuation, supporting our claim that the cross-pair construction provides a non-degenerate learning signal. Together, these results suggest that strategies which increase the proportion of hard batches—larger batch sizes and non-sentence-aware splitting—are directly beneficial for training signal quality, consistent with the ablation results in \cref{sec:ablation}.

\FloatBarrier
\Needspace{5\baselineskip}
\subsection{Sentence-aware Splitting}
\label{sec:ex_sentence_splitting}
The \emph{preserve sentence} variant in \cref{sec:ablation} constructs prefix--suffix pairs while respecting sentence boundaries and sequence-length constraints. Concretely, text documents are converted into training pairs by (i) normalizing and segmenting each document into sentence-like units and appending an end-of-sequence marker, (ii) tokenizing each unit, (iii) greedily packing consecutive units into token blocks up to a fixed maximum length, never splitting a unit except when it exceeds the maximum (then either split or discard), (iv) discarding blocks below a minimum effective length threshold, and (v) selecting within each retained block a unit boundary whose cumulative token count is closest to a desired prefix length to form a prefix (preceding tokens) and suffix (remaining tokens).

Enforcing sentence boundaries together with these length constraints has a side effect: packing and filtering discard many intermediate spans, leaving gaps between retained training examples in the original text. Consequently, examples that are adjacent in a batch can come from more distant passages and share less local context. Their cross-paired suffixes are then less related to the target prefix and easier to reject using coarse topic or context differences, weakening the pressure to learn fine-grained continuation consistency. Allowing sentence breaks retains contiguous fixed-length chunks from the text stream, preserving more of the local context shared by in-batch examples and yielding more confusable negatives. This change in batch composition helps explain the stronger gains of the \emph{break sentence} variant (\cref{fig:splitting_ablation_combo}).

\FloatBarrier
\Needspace{5\baselineskip}
\subsection{RM Training Hyperparameters}
\label{sec:ex_rm_hparams}
RM models of different backbones reported in Section \ref{sec:generalization} are trained using open source framework \texttt{verl} \citep{sheng2024hybridflow} with our proposed Algorithm \ref{alg:online_bt} and hyperparameters reported in Table \ref{tab:rm_hparams}. We perform a learning rate sweep across candidate sets $3e-7$, $5e-7$, $1e-6$, $3e-6$, and $1e-5$ and identify $1e-6$ as the best configuration that is consistent across initialization and backbone types. 

For \texttt{Qwen2.5-7B-Instruct} we use batch size $8$ which is different from the default value reported in Section \ref{sec:ablation}. This is because both instruction-tuned backbones have higher initialized performance compared to base models. In this case, scaling in-batch comparison examples provides minimal gain. Also, increasing data loader batch size under our algorithm produces quadratic increase in effective batch size, requiring more compute. Therefore, for these cases we use the smallest configuration that produces a similar performance gain.  

\begin{table}[!htbp]
\centering
\small
\caption{RM training hyperparameters.}
\label{tab:rm_hparams}
\setlength{\tabcolsep}{3pt}
\begin{tabular}{@{}l c l l c c@{}}
\toprule
\textbf{Backbone} & \textbf{Seed} & \textbf{Optimizer} & \makecell{\textbf{Learning}\\\textbf{rate}} & \makecell{\textbf{Batch}\\\textbf{size}} & \makecell{\textbf{$c$}\\\textbf{(center)}} \\
\midrule
Llama-3.2-1B & 2025 & \makecell[l]{Adam\\betas 0.9, 0.95} & \makecell[l]{$10^{-6}$, constant\\0.05 warmup ratio} & 32 & 0.01 \\
Llama-3.2-3B & 2025 & \makecell[l]{Adam\\betas 0.9, 0.95} & \makecell[l]{$10^{-6}$, constant\\0.05 warmup ratio} & 32 & 0.01 \\
Qwen2.5-3B-Inst. & 2028 & \makecell[l]{Adam\\betas 0.9, 0.95} & \makecell[l]{$10^{-6}$, constant\\0.05 warmup ratio} & 32 & 0.01 \\
Qwen2.5-7B-Inst. & 2025 & \makecell[l]{Adam\\betas 0.9, 0.95} & \makecell[l]{$10^{-6}$, constant\\0.05 warmup ratio} & 8 & 0.01 \\
\bottomrule
\end{tabular}
\end{table}

\reserveheadingroom{6\baselineskip}
\FloatBarrier
\Needspace{5\baselineskip}
\subsection{Continued-Pretraining Control}
\label{sec:ex_cpt}
We perform continued pretraining (CPT) of \modelname{Llama-3.2-3B} on the same 11M \textit{FineMath-4plus} tokens in the same order as NTP training, using the LLaMA-Factory framework \citep{zheng2024llamafactory} with the hyperparameters detailed in \cref{tab:cpt_hparams}. The two readouts in \cref{sec:ablation_controls} use 512 auxiliary perturbed preference pairs from \textit{FineMath} text blocks excluded from the earlier reward-model and continued-pretraining runs, and are evaluated on RewardBench 2. The frozen-backbone probe uses the final-layer hidden state at the last response token, the position from which our value head reads. It fits a logistic Bradley--Terry head on the difference between the chosen and rejected states, without an intercept and with the L2 penalty chosen on a 512-pair validation split. We repeat the fit on three disjoint 512-pair subsets and report the mean. Conversion jointly finetunes a newly initialized scalar reward head and all backbone parameters on 512 auxiliary pairs with the objective of \cref{eq:bt_loss}, 32 pairs per step at a constant learning rate of $10^{-5}$ for three epochs. We report the mean over three runs with independent head initializations. Applied to our 11M-token NTP reward model with its original head discarded, the probe achieves 36.4 on RewardBench 2, compared with 36.2 for the original RM head (\cref{tab:cpt_control}).\par

\begin{table}[!htbp]
\centering
\small
\caption{Continued-pretraining recipe for the 11M-token control in \cref{tab:cpt_control}. Blocks per step is the global batch size.}
\label{tab:cpt_hparams}
\setlength{\tabcolsep}{4pt}
\resizebox{\textwidth}{!}{%
\begin{tabular}{@{}l l l l c c c@{}}
\toprule
\textbf{Backbone} & \textbf{Corpus} & \textbf{Optimizer} & \makecell[l]{\textbf{Learning}\\\textbf{rate}} & \makecell{\textbf{Blocks}\\\textbf{per step}} & \makecell{\textbf{Cutoff}\\\textbf{(tokens)}} & \textbf{Epochs} \\
\midrule
Llama-3.2-3B & FineMath-4plus & AdamW & $10^{-5}$ peak, cosine, warmup ratio 0.1 & 1024 & 2048 & 1 \\
\bottomrule
\end{tabular}%
}
\end{table}

\FloatBarrier
\begin{table}[H]
\centering
\caption{RewardBench 2 scores for Llama-3.2-3B before training and after continued pretraining (CPT) or NTP reward-model training (RM), each on 11M FineMath tokens.}
\label{tab:cpt_control}
\small
\setlength{\tabcolsep}{3pt}
\renewcommand{\arraystretch}{1.12}
\begin{tabularx}{\linewidth}{@{}l *{3}{>{\centering\arraybackslash}X}@{}}
\toprule
\textbf{Readout} & \textbf{Untrained} & \textbf{CPT} & \textbf{RM} \\
\midrule
Frozen probe             & 30.0 & 28.9 & 36.4 \\
Conversion (512 pairs)   & 32.3 & 31.2 & --   \\
\rowcolor{rowgray}
Original RM head         & --   & --   & 36.2 \\
\bottomrule
\end{tabularx}\par
\end{table}

\FloatBarrier
\Needspace{10\baselineskip}
\section{Curated Dataset Details for Token-Matched Comparison}
\label{sec:ex_curated_datasets}

\FloatBarrier
\Needspace{5\baselineskip}
\subsection{Dataset Construction and Token-Equivalent Pair Counts}
\label{sec:ex_curated_datasets_stats}

\cref{sec:preference_alignment} compares our method against training on curated preference datasets under a matched 11M-token budget. Here we describe how the two datasets are constructed and annotated, followed by their token statistics and the equivalent number of preference pairs under this budget.

\paragraph{Skywork-Reward-Preference-80K-v0.2.}
We use \modelname{Skywork/Skywork-Reward-Preference-80K-v0.2} \citep{liu2024skyworkrewardbagtricks}, which contains 77,016 preference pairs over chat responses. The dataset consists primarily of Magpie-series synthetic responses ranked by ArmoRM and is decontaminated against RewardBench. Thus, it provides curated prompt--response pairs with preference labels supplied primarily by a reward model.

Based on dataset statistics, mean per-entry lengths are approximately: instruction $\approx 460$ characters ($\approx 115$ tokens), chosen response $\approx 2{,}057$ characters ($\approx 514$ tokens), rejected response $\approx 2{,}337$ characters ($\approx 584$ tokens), yielding a mean of approximately $1{,}213$ tokens per preference pair (using a standard ${\sim}4$ characters per token approximation). Under our 11M-token budget, this corresponds to training on approximately $11{,}000{,}000 / 1{,}213 \approx 9{,}070$ pairs (${\approx}$9K pairs).

\paragraph{RLHFMix1.}
\texttt{RLHFMix1} \citep{dong2024rlhf} is a curated mixture of 244,216 preference pairs. The mixture uniformly samples four sources to form a balanced training set, namely UltraFeedback \citep{cui2024ultrafeedbackboostinglanguagemodels}, HH-RLHF \citep{bai2022training}, Stanford Human Preferences (SHP) \citep{pmlr-v162-ethayarajh22a}, and Summarize-from-Feedback \citep{stiennon2022learningsummarizehumanfeedback}. UltraFeedback supplies GPT-4-judged preferences, while the other three sources provide human preferences. Thus, this baseline relies on both curated prompt--response pairs and external preference annotations.

Based on dataset statistics, mean per-entry lengths are approximately: instruction $\approx 896$ characters ($\approx 224$ tokens), chosen response $\approx 594$ characters ($\approx 149$ tokens), rejected response $\approx 481$ characters ($\approx 120$ tokens), yielding a mean of approximately $493$ tokens per preference pair. Under our 11M-token budget, this corresponds to training on approximately $11{,}000{,}000 / 493 \approx 22{,}310$ pairs (${\approx}$22K pairs).

These token-per-pair estimates indicate that our 11M-token budget is equivalent to training on roughly 9K pairs from Skywork-80K and 22K pairs from RLHFMix1—both a small fraction of each dataset's full size—yielding a comparison across supervision regimes at an approximately matched source-token budget.

\paragraph{Token budget versus compute.} Our core claim concerns data efficiency: NTP learns useful reward functions from a modest amount of existing web text without candidate generation or preference annotation. The 11M tokens used here are inexpensive to sample from existing web corpora containing trillions of tokens \citep{penedo2024fineweb}. Simple online splitting and in-batch recombination then turn this text into dense preference comparisons at little additional data-construction cost. Constructing a comparable volume of supervised preference data requires sourcing or generating candidate responses and obtaining human or model judgments, which incur substantial costs at scale (\cref{sec:ex_cost}). This advantage in data construction comes with a training-compute limitation. The budget in \cref{tab:curated_comparison} matches unique source tokens, not total computation: NTP scores all $B\times B$ prefix--continuation pairs per batch (\cref{alg:online_bt}), processing each prefix and continuation $B$ times per update. An 11M-token NTP run therefore entails roughly $B$ times as many token-forward-passes as training on 11M tokens of curated preference pairs or the continued-pretraining control (\cref{tab:cpt_control}). Our results thus support learning effective rewards from cheaply obtained web text and reducing reliance on curated supervision; they do not establish greater training-compute efficiency, nor does our central claim require it.

\begin{table}[!htbp]
\centering
\small
\caption{Per-entry token statistics for curated comparison datasets and the equivalent number of preference pairs under an 11M-token training budget. Token counts are estimated at ${\sim}4$ characters per token.}
\label{tab:curated_dataset_stats}
\setlength{\tabcolsep}{5pt}
\begin{tabular}{lrrrrrr}
\toprule
\textbf{Dataset} & \textbf{\#Pairs} & \makecell{\textbf{Instruction}\\\textbf{(tokens)}} & \makecell{\textbf{Chosen}\\\textbf{(tokens)}} & \makecell{\textbf{Rejected}\\\textbf{(tokens)}} & \makecell{\textbf{Tokens/}\\\textbf{pair}} & \makecell{\textbf{Equiv.\ pairs}\\\textbf{@ 11M tok.}} \\
\midrule
Skywork-80K-v0.2 & 77{,}016 & $\approx$115 & $\approx$514 & $\approx$584 & $\approx$1{,}213 & $\approx$9{,}070 \\
RLHFMix1         & 244{,}216 & $\approx$224 & $\approx$149 & $\approx$120 & $\approx$493   & $\approx$22{,}310 \\
\bottomrule
\end{tabular}
\end{table}

\FloatBarrier
\Needspace{5\baselineskip}
\section{Extended Details on BoN}
\label{sec:ex_bon}
\FloatBarrier
\Needspace{5\baselineskip}
\subsection{BoN Generation and Evaluation Hyperparameters}
Response generation for BoN is adapted from the open-source framework \texttt{lm-evaluation-harness} \citep{eval-harness}. We use the datasets, prompts, actor generation settings, and evaluation metrics specified in \cref{tab:bon_task_details}. For HumanEval+, we use version 0.1.10 and retain the 160 of 164 problems whose canonical solutions pass under EvalPlus. A selected program is correct only if it passes both the base and the additional test suites \citep{chen2021evaluatinglargelanguagemodels,liu2023evalplus}. Alongside the two Skywork Reward V2 baselines, we evaluate PKU's \modelname{beaver-7b-unified-reward} \citep{dai2024saferlhf,ji2025pkusaferlhfmultilevelsafetyalignment} on the same actor-generated candidate sets.

\providecommand{\bontaskcolumnwidth}{1.7cm}
\begin{table}[!htbp]
\centering
\small
\setlength{\tabcolsep}{3pt}
\renewcommand{\arraystretch}{1.15}
\newcolumntype{Y}{>{\raggedright\arraybackslash\hspace{0pt}}X}
\caption{BoN response generation and evaluation details by task.}
\label{tab:bon_task_details}
\begin{tabularx}{0.98\linewidth}{>{\raggedright\arraybackslash}p{\bontaskcolumnwidth} Y Y Y Y}
\toprule
\textbf{Task (ID/OOD)} &
\textbf{Dataset / split} &
\textbf{Prompt format} &
\textbf{Actor decoding} &
\makecell[l]{\textbf{Accuracy /}\\\textbf{scoring criterion}} \\
\midrule
MATH500 (ID) &
HuggingFaceH4/\linebreak MATH-500 (test) &
Problem: \texttt{question}. Provide your reasoning step by step, then the final answer the form FINAL\_ANSWER: \boxed{\texttt{answer}}. &
\makecell[l]{$T{=}$1.0, top-$p{=}$1.0\\max tokens${=}$512} &
\texttt{lm-eval-harness math-verify} \\
\addlinespace
GSM8K (ID) &
openai/gsm8k (test) &
Problem: \texttt{question}. Provide your reasoning step by step, then the final answer the form FINAL\_ANSWER: \boxed{\texttt{answer}}. &
\makecell[l]{$T{=}$1.0, top-$p{=}$1.0\\max tokens${=}$512} &
\texttt{lm-eval-harness math-verify} \\
\addlinespace
HumanEval+ (OOD) &
evalplus/\linebreak humanevalplus v0.1.10 (160 of 164 problems whose canonical solutions pass under EvalPlus) &
Please provide a self-contained Python script that solves the following problem in a markdown code block: \texttt{problem} &
\makecell[l]{$T{=}$1.0, top-$p{=}$1.0\\max tokens${=}$512} &
\texttt{evalplus}; pass iff the extracted program passes both the base and the plus test suites \\
\addlinespace
ToxiGen (OOD) &
toxigen/toxigen-data (test) &
Is the following statement hateful? Statement: \texttt{text}. Respond with either Yes or No. Response: &
\makecell[l]{$T{=}$1.0, top-$p{=}$1.0\\max tokens${=}$20} &
\texttt{lm-eval-harness} \\
\addlinespace
IFEval (OOD) &
google/IFEval (train) &
\texttt{prompt} &
\makecell[l]{$T{=}$1.0, top-$p{=}$1.0\\max tok.${=}$1280} &
\texttt{lm-eval-harness} prompt level strict \\
\bottomrule
\end{tabularx}
\end{table}

\FloatBarrier
\Needspace{28\baselineskip}
\section{Extended Results for RM Training}
\label{sec:ex_rm}
\FloatBarrier
\Needspace{5\baselineskip}
\subsection{Validation Reward Curve}
\label{sec:ex_reward_curve}
In Section \ref{sec:algorithm} and Section \ref{sec:ablation}, we motivate the addition of centering loss by claiming that it constrains reward magnitudes and reduces uncontrolled scale drift. Figure \ref{fig:raw_reward_curve} shows the raw positive and total reward scores on the validation set for with and without centering training configs. For both cases, raw reward difference shows an increasing trend and gets stabilized around 1.5-2 million tokens but the with centering loss case constrains the reward difference to much smaller range and maintains total reward close to zero. To provide some intuition of this centering regularization for our RM training case: we want the model to be able to learn good continuation signals from the raw text, i.e. reliably separating chosen and rejected samples, but we in general don't want to give scores of large magnitude to inherently noisy data. In fact, when running a high-quality of-the-shelf RM \texttt{Skywork-Reward-V2-Llama-3.1-8B} on an example subset of our data, we observe that it is able to score chosen and rejected samples on average differently but with a difference margin of also around 4 which resembles the margin we see in the centering loss training case. 
\begin{figure}[!htbp]
  \centering
  \includegraphics[width=0.95\textwidth]{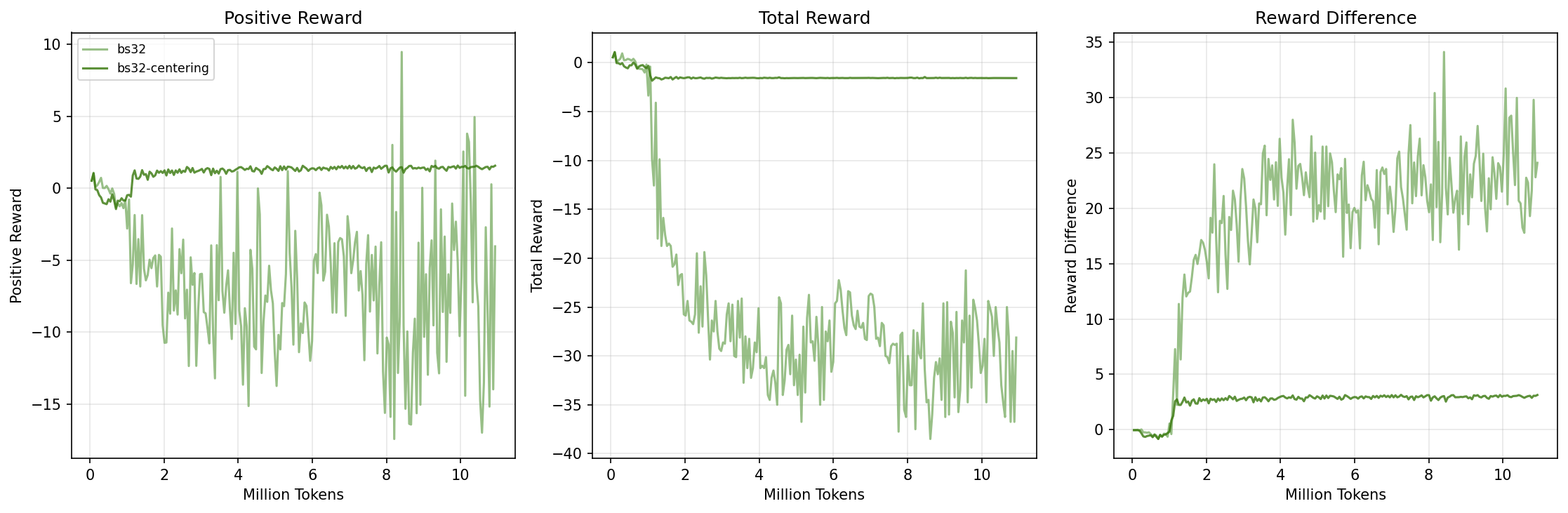}
  \caption{Validation reward curve with or without centering loss.}
  \label{fig:raw_reward_curve}
\end{figure}

\FloatBarrier
\Needspace{5\baselineskip}
\subsection{Multi-Seed Robustness}
\label{sec:ex_multiseed}

Random initialization affects the starting accuracy of an RM, consistent with previous reports \citep{liu2025skyworkrewardv2scalingpreferencedata}. Evaluating five initialization seeds for \texttt{Llama-3.2-3B} on RewardBench 2 gives a mean initial accuracy of approximately 0.26, with a 95\% confidence interval of [0.223, 0.297]; the initialization used in our main experiments falls within this interval. This variability motivates measuring improvement relative to each run's own initialization. Accordingly, our RewardBench and RewardBench 2, BoN, and actor-training evaluations include initialized RM baselines to distinguish the contribution of training from the starting performance.\par

To assess whether the gains from training are robust to seed choice, we run the default setup (\texttt{Llama-3.2-3B}, \textit{FineMath-4plus}, 11M-token budget) under three independently sampled seeds and report RewardBench 2 gains over each run's corresponding initialization in \cref{tab:multiseed}. All three runs yield positive overall gains within a narrow range of $+7.2$ to $+8.4$ percentage points, with a mean of $+7.7 \pm 0.35$ (standard error). The main-paper seed (2025) yields a representative gain of $+7.7$ points. Thus, despite variation in starting accuracy, the consistent improvements over the corresponding initializations support a reproducible learning effect from NTP training. Both overall and in-domain gains are consistent across seeds, with ID gains ranging from $+16.1$ to $+19.1$ points. Greater variability appears in how OOD gains are distributed between safety and the remaining subsets.\par

\begin{table}[!htbp]
\centering
\small
\caption{RewardBench 2 gains in percentage points over each run's initialization for three independent runs of the default \texttt{Llama-3.2-3B} setup. The final row reports means across seeds; the overall gain is reported as mean $\pm$ standard error.}
\label{tab:multiseed}
\setlength{\tabcolsep}{5pt}
\begin{tabular}{lrrrr}
\toprule
\textbf{Seed} & \textbf{ID} & \textbf{OOD Safety} & \textbf{OOD Others} & \textbf{RewardBench 2 Avg} \\
\midrule
2025 & $+16.1$ & $+5.4$ & $+7.4$ & $+7.7$ \\
2027 & $+17.5$ & $+10.5$ & $+4.7$ & $+8.4$ \\
2028 & $+19.1$ & $+16.1$ & $+0.2$ & $+7.2$ \\
\midrule
Mean & $+17.6$ & $+10.7$ & $+4.1$ & $+7.7 \pm 0.35$ \\
\bottomrule
\end{tabular}
\end{table}

\FloatBarrier
\Needspace{22\baselineskip}
\subsection[RewardBench and RewardBench 2 Per-Subset Gains]{RewardBench and RewardBench 2 Per-Subset Gains}
\label{sec:ex_rb_subsets}

To clarify which preference dimensions improve beyond the grouped results in \cref{fig:3b_rb1_rb2}, \cref{tab:rb_subset_gains} reports per-subset gains for the default \modelname{Llama-3.2-3B} setup trained on 11M tokens of \textit{FineMath-4plus}. Each value is the mean improvement over the corresponding initialization across three seeds detailed in \cref{tab:multiseed}.\par

\begin{table}[!htbp]
\centering
\caption{RewardBench and RewardBench 2 subset gains in percentage points over each run's initialization, averaged across three seeds for NTP training of \modelname{Llama-3.2-3B} on 11M FineMath tokens.}
\label{tab:rb_subset_gains}
\small
\setlength{\tabcolsep}{4pt}
\begin{tabular*}{0.8\linewidth}{@{\extracolsep{\fill}}llr@{}}
\toprule
\textbf{Benchmark} & \textbf{Subset} & \textbf{Gain} \\
\midrule
RewardBench & \texttt{Chat} & $+4.9$ \\
      & \texttt{Chat Hard} & $+7.0$ \\
      & \texttt{Safety} & $+16.0$ \\
      & \texttt{Reasoning} & $+13.1$ \\
\midrule
RewardBench 2 & \texttt{Math} & $+17.6$ \\
      & \texttt{Safety} & $+10.7$ \\
      & \texttt{Factuality} & $+2.9$ \\
      & \texttt{Ties} & $+2.7$ \\
      & \texttt{Focus} & $+10.1$ \\
      & \texttt{Precise IF} & $+2.2$ \\
\bottomrule
\end{tabular*}
\end{table}
\noindent On RewardBench, the largest gain is on \texttt{Safety} ($+16.0$), exceeding the gain on the in-domain \texttt{Reasoning} subset ($+13.1$). Improvements on \texttt{Chat} ($+4.9$) and \texttt{Chat Hard} ($+7.0$) further indicate transfer to response helpfulness and conversational preferences. On RewardBench 2, \texttt{Math} improves most ($+17.6$), but \texttt{Safety} ($+10.7$) and \texttt{Focus} ($+10.1$) also show substantial gains. These results indicate that NTP training on math-related text improves judgments of appropriate responses to harmful and benign but potentially risky requests, as well as relevance to the user's request, despite receiving no explicit preference labels for these capabilities.

Transfer is uneven, with on average smaller gains on \texttt{Factuality} ($+2.9$), \texttt{Ties} ($+2.7$), and \texttt{Precise IF} ($+2.2$). Thus, the subset breakdown supports a reward signal that transfers across several response-quality dimensions, with particularly strong improvements in safety and relevance beyond mathematical reasoning; it does not imply equally strong gains on every preference dimension.\par
\FloatBarrier
\Needspace{5\baselineskip}
\subsection[Safety Transfer Beyond Refusal Wording]{Safety Transfer Beyond Refusal Wording}
\label{sec:ex_safety_transfer}

The safety gains in \cref{sec:preference_alignment,sec:ex_rb_subsets} raise the question of whether NTP improves judgments of response content or mainly rewards superficial refusal cues. We examine this distinction using a category-level breakdown and a diagnostic subset of RewardBench 2 \citep{malik2025rewardbench2advancingreward}.\par

\emph{Category-level breakdown.} Using the CoCoNot-style taxonomy of RewardBench 2, approximately 85\% of the observed \texttt{Safety} gain comes from privacy violations, misinformation, false presuppositions, and copyright violations. These categories concern the content and implications of a response, but their contribution alone does not rule out reliance on stylistic cues. We therefore also examine examples in which refusal wording is misleading.\par

\emph{Misleading refusal wording.} For the seed-2027 \modelname{Llama-3.2-3B} run trained on 11M \textit{FineMath-4plus} tokens, the overall \texttt{Safety} gain is 10.5 percentage points over initialization (\cref{tab:multiseed}). Within this evaluation, we identify 43 ``fake-refusal'' examples whose rejected response begins with refusal boilerplate but ultimately complies with the harmful request. The model improves by 8.9 points on these examples, compared with 10.7 points on the remaining 407 examples, with both gains measured against the same run's initialization. Thus, the safety improvement persists when the presence of refusal language alone would favor the rejected response.\par

These results provide evidence against a purely stylistic explanation of safety transfer. One possible explanation is that continuation discrimination makes semantic distinctions in prompt--response compatibility more accessible to the reward function. The analysis does not establish this mechanism or rule out a contribution from style: the fake-refusal result covers a small subset from one seed and should be interpreted as diagnostic evidence.\par
\FloatBarrier
\Needspace{5\baselineskip}
\subsection[Transfer to Tool-Use Preference Evaluation]{Transfer to Tool-Use Preference Evaluation}
\label{sec:ex_tool_use}

To probe whether the learned reward signal transfers to judging tool use, we evaluate \modelname{FineMath-RM-Llama-3.2-3B} on FC-RewardBench \citep{agarwal2025toolrm}. The benchmark contains 1,500 examples derived from the single-turn splits of BFCL-v3, each providing a user query, a catalog of available tools, and correct and incorrect function-call outputs. Pairwise accuracy measures how often the reward model assigns a higher score to the correct output.\par

Our NTP-trained 3B reward model achieves 68.7\% accuracy, compared with 75.9\% for the size-matched \modelname{Skywork-Reward-V2-Llama-3.2-3B}. This provides evidence of transfer to tool-use preference judgment despite training on mathematical web text without explicit tool-use preference supervision. Both models remain substantially below the specialized ToolRM-7B and ToolRM-14B models, which achieve approximately 90\% in \citet{agarwal2025toolrm}. The remaining gap highlights the benefit of training on tool-use preference data, although those specialized models also have larger backbones.\par

This evaluation concerns preference judgments over function-call outputs from single-turn inputs. It does not establish the ability to evaluate interactive, long-horizon agent trajectories. Extending NTP to trajectory-structured corpora and evaluating rewards for multi-step agent behavior remain directions for future work.\par
\FloatBarrier
\Needspace{16\baselineskip}
\subsection{Size-matched Reward Model Ranking}
\label{sec:ex_rm_ranking}
\Cref{tab:rb_averages} summarizes the trained RMs' RewardBench and RewardBench 2 scores and rankings discussed in \cref{sec:generalization}. The full comparison sets, including our trained RMs, are listed in \cref{tab:models-0p5-1p5b,tab:models-3b-4b,tab:models-7b,tab:rb2models-0p5-1p5b,tab:rb2models-3b-4b,tab:rb2models-7b}. For the 3B backbones, the RewardBench 2 comparison includes smaller models up to a 4B parameter limit, as shown in \cref{tab:rb2models-3b-4b}.

\begin{table}[H]
\normalcolor
\centering
\caption{Average RewardBench and RewardBench 2 scores and leaderboard ranks for RMs trained with different initialization backbones on 11M tokens. Ranks are reported as $k$ of $n$, where $n$ counts all entries in the corresponding comparison set, including our models.}
\label{tab:rb_averages}

\small
\setlength{\tabcolsep}{3pt}
\renewcommand{\arraystretch}{1.12}

\begin{tabular*}{\linewidth}{@{\extracolsep{\fill}}lcccc@{}}
\toprule
\multirow{2}{*}{\textbf{Backbone}} &
\multicolumn{2}{c}{\textbf{RewardBench}} &
\multicolumn{2}{c}{\textbf{RewardBench 2}} \\
\cmidrule(lr){2-3}\cmidrule(lr){4-5}
& \textbf{Score} & \textbf{Rank} & \textbf{Score} & \textbf{Rank} \\
\midrule
Llama-3.2-1B       & 60.0 & 3 of 4   & 33.2 & 3 of 4 \\
Llama-3.2-3B       & 65.6 & 4 of 9   & 36.2 & 9 of 12 \\
Qwen2.5-3B-Inst.   & 70.0 & 3 of 9   & 46.2 & 8 of 12 \\
Qwen2.5-7B-Inst.   & 73.8 & 18 of 47 & 57.0 & 5 of 11 \\
\bottomrule
\end{tabular*}
\end{table}

\begin{table}[H]
\centering
\small
\caption{RewardBench leaderboard RMs in the 0.5--1.5B parameter range, sorted by AvgAcc.}
\label{tab:models-0p5-1p5b}
\setlength{\tabcolsep}{3pt}
\begin{tabularx}{\linewidth}{@{}>{\raggedright\arraybackslash}X r r l@{}}
\toprule
Model & Params (B) & AvgAcc & Type \\
\midrule
opencompass/CompassJudger-1-1.5B-Instruct & 1.5 & 73.4 & Generative \\
OpenAssistant/oasst-rm-2.1-pythia-1.4b-epoch-2.5 & 1.4 & 69.5 & Seq.\ Classifier \\
\rowcolor{gray!20} Finemath-RM-Llama-3.2-1B (ours) & 1.0 & 60.0 & Seq.\ Classifier \\
Qwen/Qwen1.5-0.5B-Chat & 0.5 & 53.8 & DPO \\
\bottomrule
\end{tabularx}
\end{table}

\begin{table}[H]
\centering
\small
\caption{RewardBench 2 leaderboard RMs in the 0.5--1.5B parameter range, sorted by AvgAcc.}
\label{tab:rb2models-0p5-1p5b}
\setlength{\tabcolsep}{3pt}
\begin{tabularx}{\linewidth}{@{}>{\raggedright\arraybackslash}X r r l@{}}
\toprule
Model & Params (B) & AvgAcc & Type \\
\midrule
Skywork/Skywork-Reward-V2-Llama-3.2-1B & 1.0 & 64.376 & Seq. Classifier \\
Skywork/Skywork-Reward-V2-Qwen3-0.6B & 0.6 & 61.250 & Seq. Classifier \\
\rowcolor{gray!20} Finemath-RM-Llama-3.2-1B (ours) & 1.0 & 33.2 & Seq.\ Classifier \\
OpenAssistant/oasst-rm-2.1-pythia-1.4b-epoch-2.5 & 1.4 & 26.479 & Seq. Classifier \\
\bottomrule
\end{tabularx}
\end{table}

\begin{table}[H]
\centering
\small
\caption{RewardBench leaderboard RMs in the 3-4B parameter range, sorted by AvgAcc.}
\label{tab:models-3b-4b}
\setlength{\tabcolsep}{3pt}
\begin{tabularx}{\linewidth}{@{}>{\raggedright\arraybackslash}X r r l@{}}
\toprule
Model & Params (B) & AvgAcc & Type \\
\midrule
Ray2333/GRM-llama3.2-3B-rewardmodel-ft & 3.0 & 90.9 & Seq.\ Classifier \\
stabilityai/stablelm-zephyr-3b & 3.0 & 74.0 & DPO \\
\rowcolor{gray!20} Finemath-RM-Qwen-2.5-3B (ours) & 3.0 & 70.0 & Seq.\ Classifier \\
\rowcolor{gray!20} Finemath-RM-Llama-3.2-3B (ours) & 3.0 & 65.6 & Seq.\ Classifier \\
stabilityai/stable-code-instruct-3b & 3.0 & 64.3 & DPO \\
Qwen/Qwen1.5-4B-Chat & 4.0 & 56.0 & DPO \\
ContextualAI/archangel\_sft-kto\_pythia1-4b & 4.0 & 55.9 & DPO \\
ContextualAI/archangel\_sft-dpo\_pythia1-4b & 4.0 & 52.1 & DPO \\
weqweasdas/hh\_rlhf\_rm\_open\_llama\_3b & 3.0 & 48.4 & Seq.\ Classifier \\
\bottomrule
\end{tabularx}
\end{table}

\begin{table}[H]
\centering
\small
\caption{RewardBench 2 leaderboard RMs up to 4B parameters, sorted by AvgAcc.}
\label{tab:rb2models-3b-4b}
\setlength{\tabcolsep}{3pt}
\begin{tabularx}{\linewidth}{@{}>{\raggedright\arraybackslash}X r r l@{}}
\toprule
Model & Params (B) & Avg Acc & Type \\
\midrule
Skywork/Skywork-Reward-V2-Qwen3-4B & 4.0 & 75.51 & Seq. Classifier \\
Skywork/Skywork-Reward-V2-Llama-3.2-3B & 3.0 & 74.665 & Seq. Classifier \\
Schrieffer/Llama-SARM-4B & 4.0 & 73.793 & Seq. Classifier \\
Skywork/Skywork-Reward-V2-Qwen3-1.7B & 1.7 & 68.176 & Seq. Classifier \\
Skywork/Skywork-Reward-V2-Llama-3.2-1B & 1.0 & 64.376 & Seq. Classifier \\
Skywork/Skywork-Reward-V2-Qwen3-0.6B & 0.6 & 61.25 & Seq. Classifier \\
Ray2333/GRM-gemma2-2B-rewardmodel-ft & 2.0 & 59.661 & Seq. Classifier \\
\rowcolor{gray!20} Finemath-RM-Qwen-2.5-3B (ours) & 3.0 & 46.2 & Seq.\ Classifier \\
\rowcolor{gray!20} Finemath-RM-Llama-3.2-3B (ours) & 3.0 & 36.2 & Seq.\ Classifier \\
weqweasdas/RM-Gemma-2B & 2.0 & 30.566 & Seq. Classifier \\
OpenAssistant/oasst-rm-2.1-pythia-1.4b-epoch-2.5 & 1.4 & 26.479 & Seq. Classifier \\
weqweasdas/hh\_rlhf\_rm\_open\_llama\_3b & 3.0 & 24.98 & Seq. Classifier \\
\bottomrule
\end{tabularx}
\end{table}

\begingroup
\small
\setlength{\tabcolsep}{3pt}
\sloppy
\newcommand{\ltRaggedRight}{\rightskip=0pt plus 2em\relax}
\begin{longtable}{@{}>{\ltRaggedRight\arraybackslash}p{8.0cm} r r l@{}}
\caption{RewardBench leaderboard RMs of 7B parameter size, sorted by AvgAcc.}\label{tab:models-7b} \\
\toprule
Model & Params (B) & AvgAcc & Type \\
\midrule
\endfirsthead
\multicolumn{4}{l}{\small\tablename~\thetable{} (continued)} \\
\toprule
Model & Params (B) & AvgAcc & Type \\
\midrule
\endhead
\midrule
\multicolumn{4}{r}{\small Continued on next page} \\
\endfoot
\bottomrule
\endlastfoot
Skywork/Skywork-VL-Reward-7B & 7.0 & 90.070 & Seq. Classifier \\
R-I-S-E/RISE-Judge-Qwen2.5-7B & 7.0 & 88.191 & Generative \\
internlm/internlm2-7b-reward & 7.0 & 87.593 & Seq. Classifier \\
ZiyiYe/Con-J-Qwen2-7B & 7.0 & 87.120 & Generative \\
opencompass/CompassJudger-1-7B-Instruct & 7.0 & 83.167 & Generative \\
CIR-AMS/BTRM\_Qwen2\_7b\_0613 & 7.0 & 83.152 & Seq. Classifier \\
openbmb/Eurus-RM-7b & 7.0 & 82.823 & Seq. Classifier \\
weqweasdas/RM-Mistral-7B & 7.0 & 80.389 & Seq. Classifier \\
Ahjeong/MMPO\_Gemma\_7b\_gamma1.1\_epoch3 & 7.0 & 77.444 & DPO \\
NousResearch/Nous-Hermes-2-Mistral-7B-DPO & 7.0 & 77.222 & DPO \\
Ray2333/reward-model-Mistral-7B-instruct-Unified-Feedback & 7.0 & 76.896 & Seq. Classifier \\
0-hero/Matter-0.1-7B-boost-DPO-preview & 7.0 & 76.831 & DPO \\
Ahjeong/MMPO\_Gemma\_7b & 7.0 & 76.810 & DPO \\
HuggingFaceH4/zephyr-7b-alpha & 7.0 & 76.470 & DPO \\
HuggingFaceH4/zephyr-7b-beta & 7.0 & 75.385 & DPO \\
allenai/tulu-2-dpo-7b & 7.0 & 75.163 & DPO \\
0-hero/Matter-0.1-7B-DPO-preview & 7.0 & 74.847 & DPO \\
\rowcolor{gray!20} Finemath-RM-Qwen-2.5-7B (ours) & 7.0 & 73.8 & Seq.\ Classifier \\
prometheus-eval/prometheus-7b-v2.0 & 7.0 & 72.043 & Generative \\
berkeley-nest/Starling-RM-7B-alpha & 7.0 & 71.529 & Seq. Classifier \\
ai2/tulu-2-7b-rm-v0-nectar-binarized-700k.json & 7.0 & 71.275 & Seq. Classifier \\
openbmb/Eurus-7b-kto & 7.0 & 71.048 & DPO \\
ai2/tulu-2-7b-rm-v0-nectar-binarized-3.8m-checkpoint-380k.json & 7.0 & 70.584 & Seq. Classifier \\
Qwen/Qwen1.5-7B-Chat & 7.0 & 70.579 & DPO \\
ai2/tulu-2-7b-rm-v0-nectar-binarized-3.8m-checkpoint-2660k.json & 7.0 & 70.193 & Seq. Classifier \\
ai2/tulu-2-7b-rm-v0-nectar-binarized-3.8m-checkpoint-3420k.json & 7.0 & 70.079 & Seq. Classifier \\
ai2/tulu-2-7b-rm-v0-nectar-binarized-3.8m-checkpoint-3.8m.json & 7.0 & 70.037 & Seq. Classifier \\
HuggingFaceH4/zephyr-7b-gemma-v0.1 & 7.0 & 69.562 & DPO \\
weqweasdas/RM-Gemma-7B & 7.0 & 69.542 & Seq.\ Classifier \\
ai2/tulu-2-7b-rm-v0-nectar-binarized-3.8m-checkpoint-3040k.json & 7.0 & 69.450 & Seq.\ Classifier \\
ai2/tulu-2-7b-rm-v0-nectar-binarized-3.8m-checkpoint-1900k.json & 7.0 & 69.242 & Seq.\ Classifier \\
allenai/OLMo-7B-Instruct & 7.0 & 69.216 & DPO \\
weqweasdas/RM-Gemma-7B-4096 & 7.0 & 69.095 & Seq.\ Classifier \\
ai2/tulu-2-7b-rm-v0-nectar-binarized-3.8m-checkpoint-760k.json & 7.0 & 69.045 & Seq.\ Classifier \\
ai2/tulu-2-7b-rm-v0-nectar-binarized-3.8m-checkpoint-2280k.json & 7.0 & 68.954 & Seq.\ Classifier \\
ai2/tulu-2-7b-rm-v0-nectar-binarized-3.8m-checkpoint-1140k.json & 7.0 & 68.084 & Seq.\ Classifier \\
ai2/tulu-2-7b-rm-v0-nectar-binarized.json & 7.0 & 67.558 & Seq.\ Classifier \\
RLHFlow/RewardModel-Mistral-7B-for-DPA-v1 & 7.0 & 67.038 & Seq.\ Classifier \\
ai2/tulu-2-7b-rm-v0.json & 7.0 & 66.546 & Seq.\ Classifier \\
PKU-Alignment/beaver-7b-v2.0-reward & 7.0 & 63.906 & Seq.\ Classifier \\
IDEA-CCNL/Ziya-LLaMA-7B-Reward & 7.0 & 63.681 & Seq.\ Classifier \\
PKU-Alignment/beaver-7b-v2.0-cost & 7.0 & 60.267 & Seq.\ Classifier \\
ai2/llama-2-chat-7b-nectar-3.8m.json & 7.0 & 58.427 & Seq.\ Classifier \\
PKU-Alignment/beaver-7b-v1.0-cost & 7.0 & 58.098 & Seq.\ Classifier \\
ContextualAI/archangel\_sft-kto\_llama7b & 7.0 & 53.649 & DPO \\
ContextualAI/archangel\_sft-dpo\_llama7b & 7.0 & 52.737 & DPO \\
PKU-Alignment/beaver-7b-v1.0-reward & 7.0 & 45.684 & Seq.\ Classifier \\
\end{longtable}
\endgroup

\begin{table}[H]
\centering
\small
\caption{RewardBench 2 leaderboard RMs of 7B parameter size, sorted by AvgAcc.}
\label{tab:rb2models-7b}
\setlength{\tabcolsep}{3pt}
\begin{tabularx}{\linewidth}{@{}>{\raggedright\arraybackslash}X r r l@{}}
\toprule
Model & Params (B) & Avg Acc & Type \\
\midrule
Skywork/Skywork-VL-Reward-7B & 7.0 & 68.847 & Seq. Classifier \\
weqweasdas/RM-Mistral-7B & 7.0 & 59.601 & Seq. Classifier \\
openbmb/Eurus-RM-7b & 7.0 & 58.057 & Seq. Classifier \\
CIR-AMS/BTRM\_Qwen2\_7b\_0613 & 7.0 & 57.363 & Seq. Classifier \\
\rowcolor{gray!20} Finemath-RM-Qwen-2.5-7B (ours) & 7.0 & 57.0 & Seq.\ Classifier \\
internlm/internlm2-7b-reward & 7.0 & 53.348 & Seq. Classifier \\
weqweasdas/RM-Gemma-7B & 7.0 & 48.255 & Seq. Classifier \\
PKU-Alignment/beaver-7b-v1.0-cost & 7.0 & 33.322 & Seq. Classifier \\
PKU-Alignment/beaver-7b-v2.0-cost & 7.0 & 33.261 & Seq. Classifier \\
PKU-Alignment/beaver-7b-v2.0-reward & 7.0 & 25.435 & Seq. Classifier \\
PKU-Alignment/beaver-7b-v1.0-reward & 7.0 & 16.057 & Seq. Classifier \\
\bottomrule
\end{tabularx}
\end{table}

\FloatBarrier
\Needspace{5\baselineskip}
\section{GRPO Objective and Actor-Training Details}
\label{sec:ex_actor}

\FloatBarrier
\Needspace{5\baselineskip}
\subsection{GRPO Objective}
\label{sec:grpo_objective}
For each prompt $x$, we sample $K$ completions $\{y_i\}_{i=1}^{K}\sim\pi_\theta(\cdot\mid x)$, score them with $r_\phi(x,y_i)$, and compute group-relative advantages:
\begin{equation}\label{eq:grpo_adv}
A_i=\frac{r_\phi(x,y_i)-\mu_x}{\sigma_x+\epsilon},
\end{equation}
where $\mu_x$ and $\sigma_x$ are the mean and standard deviation of the $K$ scores for $x$, and $\epsilon$ stabilizes the denominator. The actor minimizes a clipped PPO-style surrogate with reference-policy regularization:
\begin{equation}\label{eq:actor_obj}
\mathcal{L}_{\text{actor}} = \mathcal{L}_{\text{clip}}\!\left(\pi_\theta;\pi_{\theta_{\text{old}}},A\right) + \lambda\, \mathbb{E}_{t}\!\Big[ \mathcal{D}_{\mathrm{KL}}\!\big(\pi_\theta(\cdot \mid s_t)\,\|\,\pi_{\mathrm{ref}}(\cdot \mid s_t)\big) \Big].
\end{equation}
Here $\lambda$ controls deviation from the fixed reference policy, limiting policy drift under an imperfect RM. In implementation, the reference term is an MSE-style penalty on token log-probability differences \citep{schulman2017klapprox,tang2025pitfallskldivergencegradient}.

\FloatBarrier
\Needspace{5\baselineskip}
\subsection{Training Setup and Results}
\label{sec:ex_actor_results}
We train actors with GRPO using the open-source framework \texttt{verl} \citep{sheng2024hybridflow} and the hyperparameters in \cref{tab:actor_hparams}. We report the best mean@1 test accuracy reached within the five-epoch budget in \cref{tab:rm_actor_tasks}, including the \modelname{beaver-7b-unified-reward} baseline.

\begin{table}[!htbp]
\centering
\scriptsize
\caption{Actor training hyperparameters for GRPO experiments. \(\lambda\) denotes the coefficient of the reference regularization term in Eq.~(\ref{eq:actor_obj}); rollout \(K\) is the number of sampled completions per prompt.}
\label{tab:actor_hparams}

\setlength{\tabcolsep}{2pt}
\begin{tabularx}{\linewidth}{
@{}l
>{\centering\arraybackslash}p{0.6cm}
l
>{\centering\arraybackslash}p{0.8cm}
>{\centering\arraybackslash}p{0.8cm}
>{\centering\arraybackslash}p{0.8cm}
>{\centering\arraybackslash}p{0.9cm}
>{\centering\arraybackslash}p{0.9cm}
>{\raggedright\arraybackslash}X@{}
}
\toprule
\textbf{Task} &
\textbf{Ep.} &
\textbf{Optimizer} &
\makecell{\textbf{LR}} &
\makecell{\(\boldsymbol{\lambda}\)\\\textbf{KL}} &
\makecell{\textbf{Rollout}\\\(K\)} &
\makecell{\textbf{Max}\\\textbf{len.}} &
\makecell{\textbf{Batch}\\\textbf{size}} &
\makecell[l]{\textbf{Eval metric}} \\
\midrule
GSM8K & 5 & \makecell[l]{Adam\\betas 0.9, 0.95} & $10^{-6}$ & 0.1 & 8 & 512 & 512 &
mean@1 test accuracy using \texttt{math-verify} \\
MATH  & 5 & \makecell[l]{Adam\\betas 0.9, 0.95} & $10^{-6}$ & 0.1 & 8 & 512 & 512 &
mean@1 test accuracy using \texttt{math-verify} \\
\bottomrule
\end{tabularx}
\end{table}

\FloatBarrier
\Needspace{36\baselineskip}
\section{Extended Results for Continuing Training on Curated Preference Dataset}
\label{sec:ex_rlhfmix1}

Unless otherwise noted, all RMs are trained using the open source framework \texttt{Llamafactory} \citep{zheng2024llamafactory} following the hyperparameters in Table \ref{tab:rm_mid_hparams}. In Figure \ref{fig:rlhfmix1_curve}, we plot for different initialization the RewardBench 2 score of the trained RM over the 1 epoch training course on the curated preference dataset \texttt{RLHFMix1}. Results reported in \cref{tab:rewardmix1} are the RewardBench 2 performance gain achieved at the end of the training. 

\begin{table}[!htbp]
\centering
\scriptsize
\caption{Hyperparameters for RM training with curated preference dataset \texttt{RLHFMix1}. Cutoff length indicates the token number limits for prompt and response concatenated sequence.}
\label{tab:rm_mid_hparams}

\setlength{\tabcolsep}{3pt}
\begin{tabular}{
@{}l 
c 
l 
>{\centering\arraybackslash}p{0.9cm} 
>{\centering\arraybackslash}p{1.0cm} 
>{\centering\arraybackslash}p{1.1cm} 
>{\centering\arraybackslash}p{1.1cm}@{}
}
\toprule
\textbf{Dataset} &
\textbf{Epochs} &
\textbf{Optimizer} &
\textbf{LR} &
\makecell{\textbf{Warmup}\\\textbf{ratio}} &
\makecell{\textbf{Cutoff}\\\textbf{len.}} &
\makecell{\textbf{Batch}\\\textbf{size}} \\
\midrule
\texttt{RLHFMix1} & 1 & \makecell[l]{Adam\\betas 0.9, 0.95} & $2e$--$6$ & 0.03 & 2048 & 512  \\
\bottomrule
\end{tabular}
\end{table}

\begin{figure}[!htbp]
  \centering
  \includegraphics[width=0.90\textwidth]{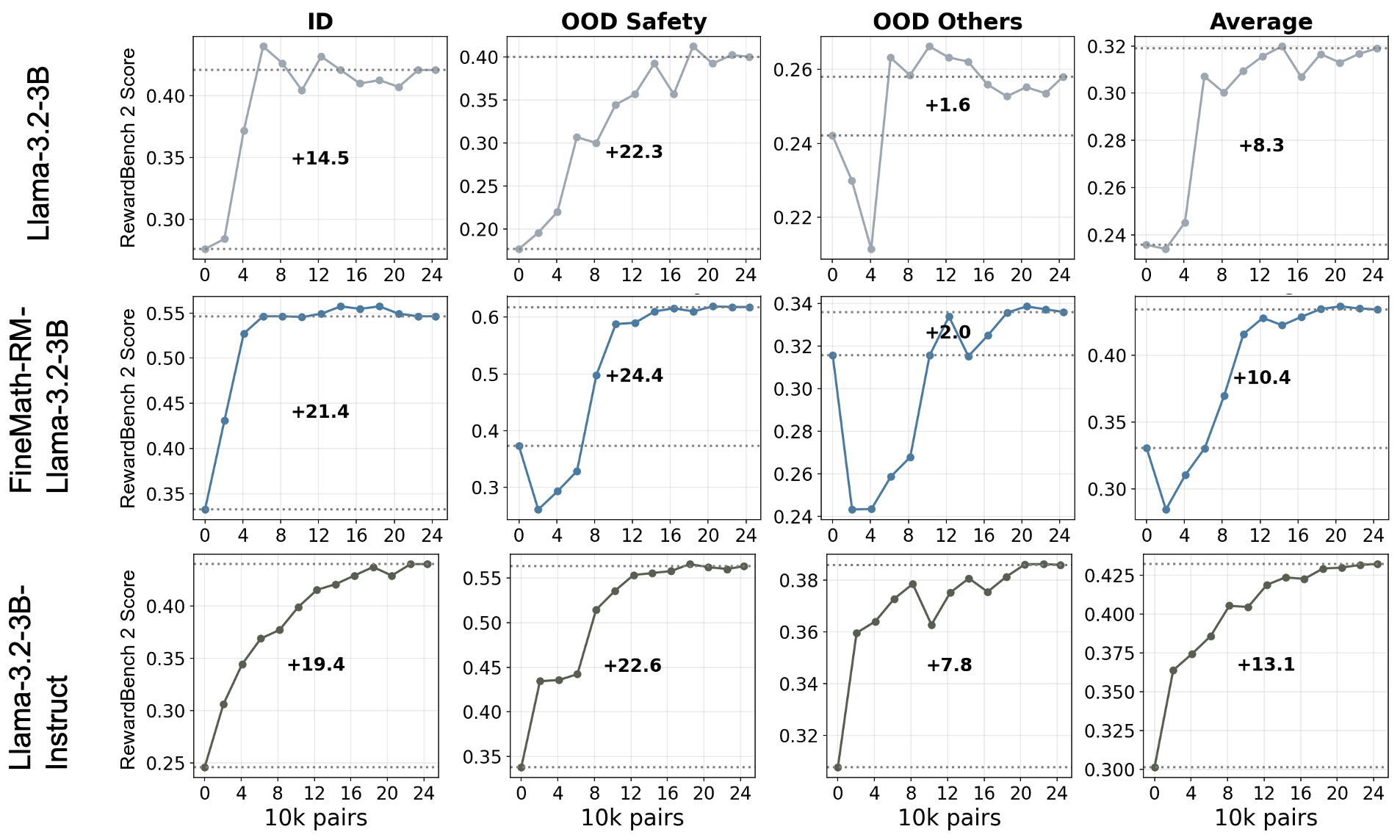}
  \captionsetup{justification=raggedright,singlelinecheck=false}
  \caption{RewardBench 2 curve across the 1 epoch training course on the curated preference dataset \texttt{RLHFMix1} for 3 representative initializations \texttt{Llama-3.2-3B} base model, \texttt{FineMath-RM-Llama-3.2-3B} trained from our method, and \texttt{Llama-3.2-3B-Instruct}.}
  \label{fig:rlhfmix1_curve}
\end{figure}

\FloatBarrier
\Needspace{28\baselineskip}
\subsection[Label Efficiency of NTP Initialization]{Label Efficiency of NTP Initialization}
\label{sec:ex_label_efficiency}

To measure how NTP initialization affects the amount of curated supervision needed for reward modeling, we compare the base \modelname{Llama-3.2-3B} checkpoint with its NTP-trained counterpart across six fractions of \texttt{RLHFMix1}. NTP initialization uses 11M tokens of \textit{FineMath-4plus} before curated preference training. We report absolute overall RewardBench 2 scores, allowing direct comparison between the two initializations at each labeled-data budget.\par

\begin{table}[!htbp]
\centering
\captionsetup{font=small,skip=3pt}
\caption{Overall RewardBench 2 scores after training on different fractions of \texttt{RLHFMix1}, comparing base and NTP initialization. }
\label{tab:label_efficiency}
\small
\setlength{\tabcolsep}{3pt}
\renewcommand{\arraystretch}{1.1}
\begin{tabular*}{0.75\linewidth}{@{\extracolsep{\fill}}rrrr@{}}
\toprule
\makecell{\textbf{Labeled}\\\textbf{fraction}} & \textbf{Pairs} & \makecell{\textbf{Base}\\\textbf{init.}} & \makecell{\textbf{NTP}\\\textbf{init.}} \\
\midrule
1\%   & 2.4K & 23 & 32 \\
5\%   & 12K  & 23 & 29 \\
10\%  & 24K  & 24 & 29 \\
25\%  & 61K  & 31 & 33 \\
50\%  & 122K & 32 & 44 \\
100\% & 244K & 32 & 44 \\
\bottomrule
\end{tabular*}
\end{table}
\noindent\Cref{tab:label_efficiency} shows higher scores from NTP initialization at every evaluated budget. With only 1\% of the labeled data (approximately 2.4K pairs), NTP initialization reaches approximately 32, compared with 23 for base initialization. At 50\% (approximately 122K pairs), NTP initialization reaches approximately 44, matching its result with the full dataset. Base initialization reaches 32 at both 50\% and 100\%, leaving a gap of about 12 points at these budgets.

Thus, NTP initialization reaches its full-data performance using half of the curated preference pairs, while exceeding the score attained by base initialization at any of the tested budgets. The benefit is not a uniform improvement as more labels are added: the NTP-initialized model's score decreases from approximately 32 at 1\% to 29 at 5--10\%, before recovering and improving at larger budgets. Nevertheless, its advantage over base initialization persists throughout the sweep. These results support using NTP training on web text to improve the label efficiency of subsequent supervised reward modeling.\par
\FloatBarrier
\Needspace{5\baselineskip}
\section{Licenses for Existing Assets}
\label{sec:licenses}

\paragraph{Datasets.}
\begin{itemize}[itemsep=1pt,parsep=0pt,topsep=3pt]
    \item \textbf{FineMath-4plus} \citep{allal2025smollm2smolgoesbig} — released by HuggingFaceFW under the ODC-By 1.0 license.
    \item \textbf{FineWeb-Edu} \citep{penedo2024fineweb} --- released by HuggingFaceFW under the ODC-By 1.0 license.
    \item \textbf{MegaMath-Web-Pro-Max} \citep{wang2025octothinkermidtrainingincentivizesreinforcement} --- the OctoThinker dataset card lists ODC-By 1.0 and specifies non-commercial use and compliance with the original source licenses.
    \item \textbf{RLHFMix1} — a mixture of UltraFeedback \citep{cui2024ultrafeedbackboostinglanguagemodels} (MIT), HH-RLHF \citep{bai2022training} (MIT), SHP \citep{pmlr-v162-ethayarajh22a} (CC BY 4.0), and Summarize from Feedback \citep{stiennon2022learningsummarizehumanfeedback} (MIT).
    \item \textbf{Skywork-Reward-Preference-80K-v0.2} \citep{liu2024skyworkrewardbagtricks} — released under CC BY 4.0.
    \item \textbf{GSM8K} \citep{cobbe2021trainingverifierssolvemath} — released by OpenAI under MIT.
    \item \textbf{MATH} \citep{hendrycks2021measuring} — released under MIT.
    \item \textbf{ToxiGen} \citep{hartvigsen2022toxigenlargescalemachinegenerateddataset} — released under MIT.
    \item \textbf{IFEval} \citep{zhou2023instructionfollowingevaluationlargelanguage} — released under Apache 2.0.
    \item \textbf{HumanEval+} \citep{chen2021evaluatinglargelanguagemodels,liu2023evalplus} --- HumanEval released by OpenAI under MIT; the EvalPlus test suites released under Apache 2.0.
    \item \textbf{RewardBench} \citep{lambert2024rewardbenchevaluatingrewardmodels} and \textbf{RewardBench 2} \citep{malik2025rewardbench2advancingreward} — released under Apache 2.0.
\end{itemize}

\Needspace{12\baselineskip}
\paragraph{Models.}
\begin{itemize}[itemsep=1pt,parsep=0pt,topsep=3pt]
    \item \textbf{Llama 3 family} (\texttt{Llama-3.2-1B}, \texttt{Llama-3.2-3B}, \texttt{Llama-3.1-8B} and Instruct variants) \citep{grattafiori2024llama3herdmodels} — released under the Llama 3 Community License.
    \item \textbf{Qwen2.5 family} (\texttt{Qwen2.5-3B-Instruct}, \texttt{Qwen2.5-7B-Instruct}) \citep{qwen2025qwen25technicalreport} — released under Apache 2.0.
    \item \textbf{Skywork-Reward-V2} models \citep{liu2025skyworkrewardv2scalingpreferencedata} — released under Apache 2.0.
    \item \textbf{\modelname{PKU-Alignment/beaver-7b-unified-reward}} \citep{dai2024saferlhf,ji2025pkusaferlhfmultilevelsafetyalignment} --- released under a non-commercial license.
\end{itemize}

\Needspace{8\baselineskip}
\paragraph{Frameworks and Tools.}
\begin{itemize}[itemsep=1pt,parsep=0pt,topsep=3pt]
    \item \textbf{verl} \citep{sheng2024hybridflow} — released under Apache 2.0.
    \item \textbf{LlamaFactory} \citep{zheng2024llamafactory} — released under Apache 2.0.
    \item \textbf{lm-evaluation-harness} \citep{eval-harness} — released under MIT.
    \item \textbf{math-verify} — released under Apache 2.0.
\end{itemize}

\FloatBarrier
\Needspace{13\baselineskip}
\section{Compute Resources}
\label{sec:compute}

All experiments are run on NVIDIA A100 or H100 80GB GPUs.
The approximate compute cost per experimental type is summarized below.

\begin{itemize}[itemsep=1pt,parsep=0pt,topsep=3pt]
    \item \textbf{RM training.} Each run uses 1, 2, or 4 H100 GPUs and takes a few hours depending on the backbone size and training budget.
    \item \textbf{CPT training.} The continued-pretraining control uses 4 H100 GPUs, processes 11M tokens in one epoch, and takes less than one hour.
    \item \textbf{Best-of-$N$ (BoN) evaluation.} Each evaluation run uses 1 H100 GPU and takes a few hours.
    \item \textbf{Actor training.} Each run uses 2 H100 GPUs and takes approximately half a day.
    \item \textbf{RM mid-training (curated preference dataset).} Each run uses 1 H100 GPU and takes a few hours.
\end{itemize}

\FloatBarrier
\endgroup

\arxivFloatBarrier

\end{document}